\documentclass{article}

 \usepackage[preprint]{neurips_2026}

\usepackage[utf8]{inputenc} 
\usepackage[T1]{fontenc}    
\usepackage{url}            
\usepackage{booktabs}       
\usepackage{amsfonts}       
\usepackage{amsmath}        
\usepackage{nicefrac}       
\usepackage{microtype}      
\usepackage{xcolor}         

\definecolor{citeblue}{RGB}{31,119,180}
\usepackage[colorlinks=true, citecolor=citeblue, linkcolor=red, urlcolor=blue]{hyperref}

\usepackage{multirow}       
\usepackage{graphicx}       
\usepackage{tikz}           
\usetikzlibrary{positioning,arrows.meta,shapes.geometric,fit,backgrounds,calc}
\usepackage{xspace}
\usepackage{cleveref}
\usepackage{colortbl}       
\usepackage{pifont}         
\usepackage{caption}        
\usepackage{float}          
\usepackage{makecell}
\usepackage{titletoc}
\usepackage{graphicx}
\usepackage{subcaption}

\usepackage{wrapfig}
\usepackage{tcolorbox}
\tcbuselibrary{breakable,skins}

\usepackage{caption}
\usepackage{marvosym}

\newcommand{\cmark}{{\color{green!50!black}\ding{51}}}
\newcommand{\xmark}{{\color{red!70!black}\ding{55}}}

\newcommand{\benchname}{\textbf{MSAVBench}\xspace}
\newcommand{\benchnameplain}{MSAVBench\xspace}

\newcommand{\eg}{\textit{e.g.}\xspace}

\definecolor{rkone}{cmyk}{0.45,0.00,0.05,0.00}   
\definecolor{rktwo}{cmyk}{0.32,0.00,0.04,0.00}   
\definecolor{rkthree}{cmyk}{0.22,0.00,0.03,0.00} 
\definecolor{rkfour}{cmyk}{0.13,0.00,0.02,0.00}  
\definecolor{rkfive}{cmyk}{0.06,0.00,0.01,0.00}  

\newcommand{\rkone}[1]{\cellcolor{rkone}{#1}}
\newcommand{\rktwo}[1]{\cellcolor{rktwo}{#1}}
\newcommand{\rkthree}[1]{\cellcolor{rkthree}{#1}}
\newcommand{\rkfour}[1]{\cellcolor{rkfour}{#1}}
\newcommand{\rkfive}[1]{\cellcolor{rkfive}{#1}}

\title{\benchname: Towards Comprehensive and Reliable Evaluation of Multi-Shot Audio-Video Generation}

\author{%
\bf
Yujie Wei$^{1}$\thanks{Equal Contribution\quad$^\dagger$Project Leader\quad$\textsuperscript{\Letter}$Corresponding Authors}~,~~Yujin Han$^{2*}$,~~Zhekai Chen$^{2*}$,~~Yongming Li$^{1*}$,~~Kaixun Jiang$^{1}$,
\\
\bf
Zhihang Liu$^3$,~~Quanhao Li$^{1}$,~~Zhiwu Qing$^3$,~~Xiang Wang$^3$,~~Zhen Xing$^3$,~~Ruihang Chu$^3$,
\\
\bf
Lingyi Hong$^{1}$,~~Yefei He$^4$,~~Junjie Zhou$^3$,~~Junqiu Yu$^{1}$,~~Yang Shi$^5$,~~Difan Zou$^{2}$,~~Kai Zhu$^3$,
\\
\bf
Shiwei Zhang$^{3\dagger}$,~~Yingya Zhang$^3$,~~Yu Liu$^3$,~~Xihui Liu$^{2\text{\Letter}}$,~~Hongming Shan$^{1\text{\Letter}}$
 \vspace{1mm}\\
 $^1$Fudan University\quad
 $^2$The University of Hong Kong\quad
 $^3$Tongyi Lab, Alibaba Group
 \\
 $^4$Zhejiang University
 \quad
 $^5$Peking University
}

\begin{document}

\maketitle

\begin{abstract}
Video generation is rapidly evolving from single-shot synthesis to complex multi-shot audio-video (MSAV) narratives to meet real-world demands. However, evaluating such frontier models remains a fundamental challenge. Existing benchmarks are limited in scope and data diversity, and rely on rigid evaluation pipelines, preventing systematic and reliable assessment of modern MSAV models.
To bridge these gaps, we introduce \benchname, the first comprehensive benchmark and adaptive hybrid evaluation framework for multi-shot audio-video generation.
Our benchmark spans four key dimensions, video, audio, shot, and reference, covering diverse task settings, varying shot counts of up to 15, and challenging non-realistic scenarios.
Our evaluation framework improves robustness through an adaptive self-correction mechanism for shot segmentation, instance-wise rubrics for subjective metrics, and tool-grounded evidence extraction for complex judgments.
Furthermore, \benchnameplain\ achieves high alignment with human judgments, reaching a Spearman rank correlation of 91.5\%.
Our systematic evaluation of 19 state-of-the-art closed- and open-source models shows that current systems still struggle with director-level control and fine-grained audio-visual synchronization, while modular or agentic generation pipelines offer a promising path toward narrowing the gap between open- and closed-source models.
The benchmark data and evaluation code are publicly available at \url{https://github.com/ali-vilab/MSAVBench}.
\end{abstract}

\section{Introduction}

The landscape of generative video is shifting from \emph{silent, single-shot} text-to-video (T2V) synthesis~\citep{sora,hunyuanvideo,ltx-video} toward \emph{multi-shot audio-video} (MSAV) generation~\citep{seedance2.0,wan2.7,sora-2}. Unlike traditional short clips, MSAV enables cinematic storytelling with complex narratives and synchronized audio.
While frontier closed-source systems (\eg, Seedance~2.0~\citep{seedance2.0}, Wan~2.7~\citep{wan2.7}, Sora~2~\citep{sora-2}) have demonstrated impressive MSAV capabilities, the open-source community currently lacks dedicated MSAV models, leaving a critical gap in the field. Therefore, establishing a comprehensive MSAV benchmark is an urgent prerequisite to providing design guidelines for the open-source community and to diagnosing model weaknesses in closed-source systems.

However, evaluating MSAV generation is fundamentally challenging due to its compositional, multi-shot, and multi-modal nature.
Specifically, existing benchmarks only address isolated facets of this problem, falling short on two concrete fronts:
\textbf{(i) Limited evaluation scope and data diversity.}
Most prior benchmarks~\citep{vbench, evalcrafter, videobench} target \emph{single-shot}, silent generation.
Recent efforts only partially bridge this gap: they focus either on single-shot audio-video generation~\citep{avgenbench}, or on multi-shot video synthesis but lack thorough audio evaluation~\citep{msvbench, openS2V, vistorybench}.
Furthermore, their evaluation datasets exhibit limited diversity and complexity, overlooking the rich cinematic language and challenging scenarios like counterfactual content.
Consequently, these benchmarks fail to systematically assess the diverse task adaptability and performance of modern MSAV models in complex scenarios.
\textbf{(ii) Rigid and static evaluation pipelines.}
First, they struggle with \textit{limited robustness to shot mis-segmentation}. 
Generated videos often exhibit variable shot counts and ambiguous transition boundaries, making shot-based evaluation highly sensitive to segmentation errors. Existing pipelines typically rely on fixed segmenters without self-correction, so a single mis-segmentation can distort downstream metrics.
Second, they employ \textit{rigid scoring paradigms for complex dimensions}. For important yet challenging dimensions without dedicated expert models (\eg, narrative coherence and layout--text consistency), existing pipelines often rely on direct VLM scoring. Although simple to implement, this strategy is sensitive to prompt phrasing and prone to hallucination, making it unreliable for assessing performance on complex tasks.

\begin{figure}[t]
    \centering
    \includegraphics[width=1\linewidth]{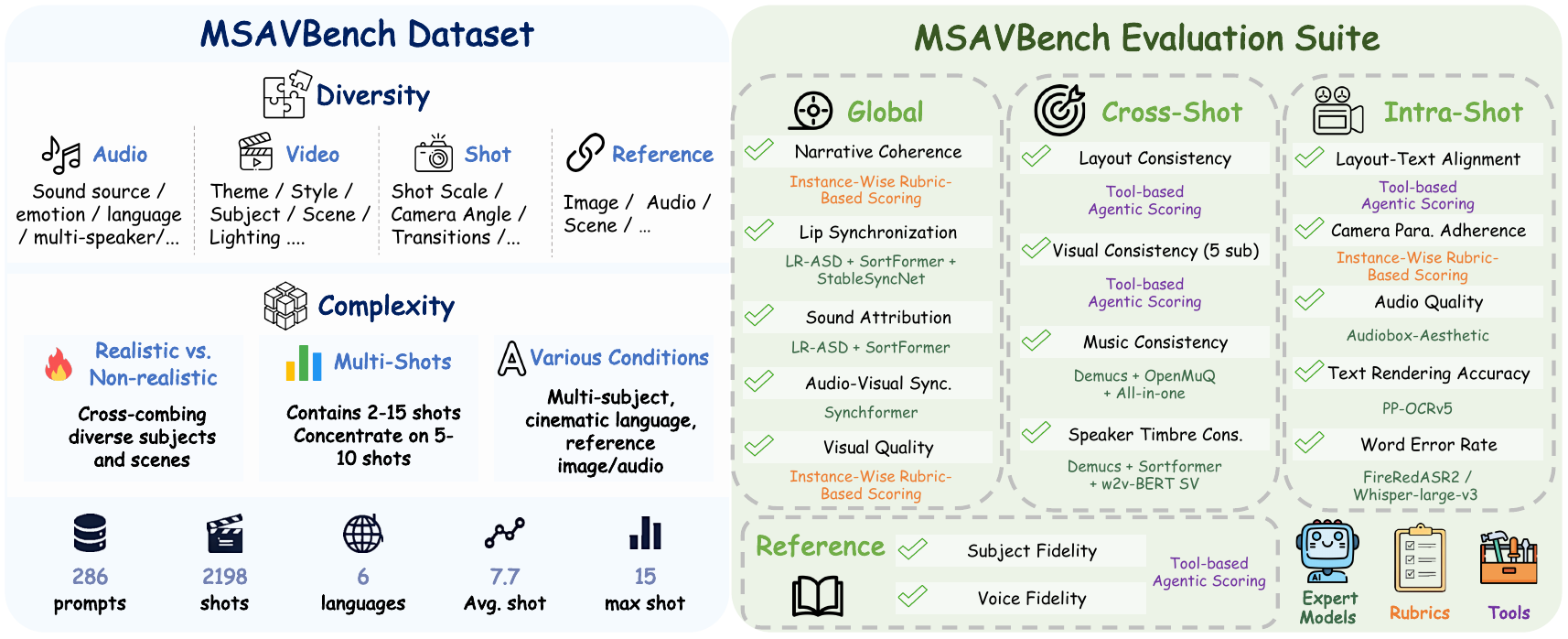}
    \caption{\textbf{Overview of \benchname.} Left: the benchmark spans four data dimensions, namely video, audio, shot, and reference, covering diverse prompts, shot counts, and realistic and non-realistic scenarios. Right: the evaluation suite assesses generated MSAV content at four levels, including global, cross-shot, intra-shot, and reference levels, using a hybrid strategy that combines specialized expert models, rubric-based scoring, and tool-grounded assessment.}
    \label{fig:teaser}
    \vspace{-5mm}
\end{figure}

To bridge these gaps, we present \benchname, a comprehensive benchmark and adaptive hybrid evaluation framework for MSAV generation, as shown in \Cref{fig:teaser}.
First, our benchmark is designed for broad and challenging coverage. It spans four key dimensions: \textit{video}, \textit{audio}, \textit{shot}, and \textit{reference}, each with diverse sub-dimensions, and includes a wide range of generation settings, such as varying shot counts (up to 15), different numbers of subjects, and non-realistic scenarios.
Second, the evaluation framework is designed for robustness and reliability. We introduce a self-correction mechanism that enables a VLM to iteratively inspect shot boundaries and invoke tools to merge or split segments, thereby mitigating error propagation from shot mis-segmentation.
For subjective dimensions such as narrative coherence, we replace direct VLM scoring with instance-wise rubrics formulated as predefined multiple-choice questions.
For complex dimensions such as layout--text consistency, we allow the model to adaptively invoke external perception tools to gather objective evidence for the final judgment.
Together, \benchnameplain\ enables a more comprehensive and reliable assessment of modern MSAV models, revealing their multifaceted capabilities and limitations while achieving high alignment with human judgments, reflected by a Spearman rank correlation of 91.5\%.

Leveraging \benchnameplain, we conduct a comprehensive evaluation of 19 state-of-the-art closed- and open-source models. Our analysis reveals three key insights into the current MSAV landscape: \textit{(i)} a substantial performance gap persists between closed- and open-source systems, but modular or agentic generation pipelines show promise for narrowing this gap; \textit{(ii)} current models remain far from reliable ``director-level'' generation, struggling with cinematic control, structural consistency, and fine-grained joint audio-visual alignment; and \textit{(iii)} the common ``video-first, post-hoc dubbing'' paradigm is insufficient for complex multi-shot audio-video generation, highlighting the need for unified audio-video architectures.

In summary, our contributions are threefold. \textit{First}, we release \benchname, the first benchmark for multi-shot audio-video generation, covering four key dimensions: video, audio, shot, and reference, as well as diverse tasks and challenging generation settings. \textit{Second}, we propose an adaptive hybrid evaluation framework that improves robustness through dynamic shot-boundary correction, instance-wise rubrics, and tool-grounded evidence extraction. \textit{Third}, we systematically evaluate 19 state-of-the-art closed- and open-source models, showing that modular and agentic generation pipelines are a promising path for open-source systems, while highlighting challenges in director-level control and audio-visual synchronization as well as the need for unified audio-video architectures.

\begin{table}[t]
\centering
\caption{\textbf{Comparison with existing video and audio--video generation benchmarks.} 
\textbf{Counterf.}: counterfactual prompts; \textbf{Cine.}: cinematic language and camera control; \textbf{Ref.}: reference conditioning. \benchnameplain offers comprehensive coverage of data dimensions and challenging cases, along with a robust evaluation framework featuring adaptive self-correction and agentic scoring.}
\label{tab:bench_comparison}
\setlength{\tabcolsep}{3.5pt}
\renewcommand{\arraystretch}{1.05}
\resizebox{\linewidth}{!}{%
\begin{tabular}{l c c | c c c c c | c c | c c}
\toprule
\textbf{Benchmark}
  & \makecell{\textbf{Avg.}\\\textbf{Shots}} 
  & \makecell{\textbf{Counterf.}}
  & \textbf{Video} 
  & \textbf{Audio} 
  & \makecell{\textbf{Audio-}\\\textbf{Video}} 
  & \textbf{Cine.} 
  & \textbf{Ref.}
  & \makecell{\textbf{Shot}\\\textbf{Correction}} 
  & \makecell{\textbf{Agentic}\\\textbf{Scoring}}
  & \makecell{\textbf{\#}\\\textbf{Metrics}} 
  & \makecell{\textbf{\#}\\\textbf{Prompts}} \\
\midrule
VBench~\citep{vbench}                  & 1        & \xmark & \cmark & \xmark & \xmark & \xmark & \xmark & \xmark & \xmark   & 16 & $\sim$1{,}600 \\
EvalCrafter~\citep{evalcrafter}        & 1        & \xmark & \cmark & \xmark & \xmark & \xmark & \xmark & \xmark & \xmark   & 17 & 700     \\
Video-Bench~\citep{videobench}         & 1        & \xmark & \cmark & \xmark & \xmark & \xmark & \xmark & \xmark & \xmark   & 9  & 419     \\
OpenS2V-Nexus~\citep{openS2V}          & 1        & \xmark & \cmark & \xmark & \xmark & \xmark & \cmark & \xmark & \xmark   & 6  & 180     \\
ViStoryBench~\citep{vistorybench}      & 16.5     & \cmark & \cmark & \xmark & \xmark & \cmark & \cmark & \xmark & \xmark   & 12 & 80      \\
MSVBench~\citep{msvbench}              & $\sim$14 & \cmark & \cmark & \xmark & \xmark & \cmark & \cmark & \xmark & \xmark   & 20 & 20      \\
UniVBench~\citep{univbench}            & 3.72     & \cmark & \cmark & \xmark & \xmark & \cmark & \cmark & \xmark & \cmark   & 21 & 200     \\
AVGen-Bench~\citep{avgenbench}         & 1.6      & \xmark & \cmark & \cmark & \cmark & \xmark & \xmark & \xmark & \xmark   & 10 & 235     \\
\midrule
\rowcolor{gray!12}
\textbf{\benchnameplain (Ours)}
  & \textbf{7.7} & \cmark
  & \cmark & \cmark & \cmark & \cmark & \cmark
  & \cmark & \cmark
  & \textbf{20} & \textbf{286} \\
\bottomrule
\end{tabular}%
\vspace{-3mm}
}
\end{table}
\section{Related Work}

\textbf{Audio-video generation models.} \quad
Building upon the success of image generation~\citep{ho2020denoising, wan-image, wei2025routing, esser2024scaling, liao2026aibench}, current video generative models mainly target single-shot video synthesis~\citep{sora,hunyuanvideo,ltx-video, singer2022make, ho2022video, wei2024dreamvideo, wei2025dreamrelation}.
While yielding impressive results, this paradigm is insufficient for scenarios requiring multi-scene narratives and synchronized audio~\citep{blattmann2023stable,polyak2024movie,wei2024dreamvideo2, wei2026dreamvideo-omni}.
More recently, frontier closed-source systems have explored multi-shot audio-video generation~\citep{sora-2,wan2.7,seedance2.0,happyhorse,kling-3,veo-3.1}, while open-source efforts remain limited and often rely on multi-shot video generation followed by audio dubbing~\citep{shotstream,longlive,helios,huang2025selfforcingbridgingtraintest,zhu2026causalforcingautoregressivediffusion,hunyuanvideofoley,cheng2025mmaudiotamingmultimodaljoint,wang2024av,zhao2025uniform,polyak2024movie,guan2025audcast}.
However, evaluation of MSAV models remains underexplored and highly challenging due to the need to assess both long-range multi-shot coherence and fine-grained audio-visual alignment.

\textbf{Audio-video evaluation benchmarks.} \quad
Early benchmarks such as VBench~\citep{vbench}, Video-Bench~\citep{videobench}, and AesVideo-Bench~\citep{han2026aesrm} mainly assess single-shot visual quality. Later multi-shot benchmarks~\citep{vistorybench,univbench,shotstream,msvbench} extend evaluation to story structure and cross-shot consistency, but remain largely video-centric with limited audio assessment.
Meanwhile, audio-video benchmarks~\citep{avgenbench,phyavbench,mtavg-bench,vabench,t2av-compass} evaluate audio quality and audio-visual alignment, yet mostly focus on single-shot or weakly structured prompts, with limited coverage of complex multi-shot settings and challenging scenarios such as counterfactual compositions.
Their evaluation pipelines are also typically static, making it difficult to reliably assess more complex dimensions.
In contrast, as summarized in \Cref{tab:bench_comparison}, \benchnameplain is tailored to multi-shot audio-video generation, combining broad coverage of data settings and challenging cases, together with a robust and adaptive evaluation framework that supports self-correction and agentic scoring.

\section{\benchnameplain}
\label{sec:method}

\subsection{Data Design}
\label{sec:method:data-def}

To comprehensively evaluate the MSAV ability of existing audio-video generation models, our data design is guided by two core dimensions: diversity and complexity.

\textbf{Diversity.}\quad We decompose the MSAV generation task into four primary dimensions to ensure broad data coverage:
\textbf{1) Video:} Spans diverse generation categories, visual styles, and subject types across varying scenes, color tones, and lighting conditions.
\textbf{2) Audio:} Encompasses a wide range of sound sources, affective states (emotions), and multilingual spoken content.
\textbf{3) Shot:} Introduces explicit professional cinematic language, including shot scales, camera angles, movement patterns, and cross-shot transitions.
\textbf{4) Reference:} Extends beyond standard text-conditioned generation by incorporating reference conditions, such as characters, scenes, and audio, to evaluate identity and timbre preservation.
A detailed distribution analysis is provided in Sec.~\ref{sec:method:data-ana}.

\textbf{Complexity.}\quad
Beyond data diversity, data complexity is essential to probe the performance limits of existing models. We structure this complexity across two main perspectives:
\textbf{1) Reality and Non-reality:} We explicitly categorize both subjects and scenes into \emph{realistic} and \emph{non-realistic} domains. The latter encompasses fictional worlds and counterfactual compositions. By cross-combining these axes, we evaluate a model's ability to faithfully adhere to complex prompts without mode collapse or falling back to common real-world data biases.
\textbf{2) Challenging Scenarios:} We include a diverse range of challenging settings across both video and audio. These include overlapping simultaneous audio sources, complex fast-paced motions, dense on-screen text rendering, and diverse languages. Most importantly, we push the structural boundaries of MSAV generation by extending narratives up to \textit{15} shots, together with varying subject counts and mixed cinematic transitions.

\subsection{Data Construction}
\label{sec:method:data-cons}

To construct a high-quality benchmark adhering to the two data design principles, we introduce a four-stage pipeline integrating automated generation with human annotation in \Cref{fig:data_construct}.

\textbf{Stage 1: Expert-driven taxonomy and quadruple construction.}\quad
Domain experts first define an 8-category taxonomy based on video content genres (detailed in Sec.~\ref{sec:method:data-ana}), which is further decomposed into fine-grained themes to prevent prompt homogenization. Concurrently, experts curate extensive candidate pools for subjects, scenes, and visual styles, strictly categorizing them into realistic and non-realistic domains. This process yields a vast combinatorial pool of $(theme, subject, scene, style)$ seed quadruples (see the Appendix~\ref{app:data_construction_details} for the complete taxonomy).

\textbf{Stage 2: Prompt generation and rewriting.}\quad
We randomly sample 2200 seed quadruples, and employ GPT-5.4~\citep{gpt-5.4} to synthesize initial prompts based on these quadruples while extracting structured evaluation metadata (\eg, shot counts, audio categories). We then use a \textit{Prompt Enhancement model}
to rewrite these initial prompts into comprehensive \emph{global-to-shot} scripts. Each structured script comprises a global overview followed by detailed per-shot captions, which are enriched with explicit cinematic language, including camera parameters, transition cues, and lighting conditions.

\textbf{Stage 3: Expert annotation and refinement.}\quad
Six domain experts rigorously review the 2200 generated scripts to ensure diversity, structural complexity, and logical coherence. Experts filter out redundant and homogeneous cases, unnatural cross-shot transitions, and LLM hallucinations (\eg, semantic deviations from the initial scripts), while manually refining ambiguous descriptions. This strict curation yields a high-quality prompt suite of 286 prompts comprising 2198 individual shots.

\textbf{Stage 4: Reference media collection.}\quad
To support reference-conditioned generation, we first sample 1000 character image-audio pairs (spanning both realistic and anime domains) and 200 background images from established public benchmarks~\citep{talkvid,anim-400k,univbench,anime-scene}. Next, we use a VLM (Gemini 3.1 Pro~\citep{gemini-3.1-pro}) to categorize these assets to align with the semantic conditions of our scripts. We then enforce strict global uniqueness constraints to map these candidates to specific scripts, while human experts meticulously filter out low-quality samples or misaligned matches. This yields a reliable reference subset of 68 subject images, 65 audio clips, and 32 scene images, assigned across 96 scripts.

\begin{figure}[t]
    \centering
    \includegraphics[width=1\linewidth]{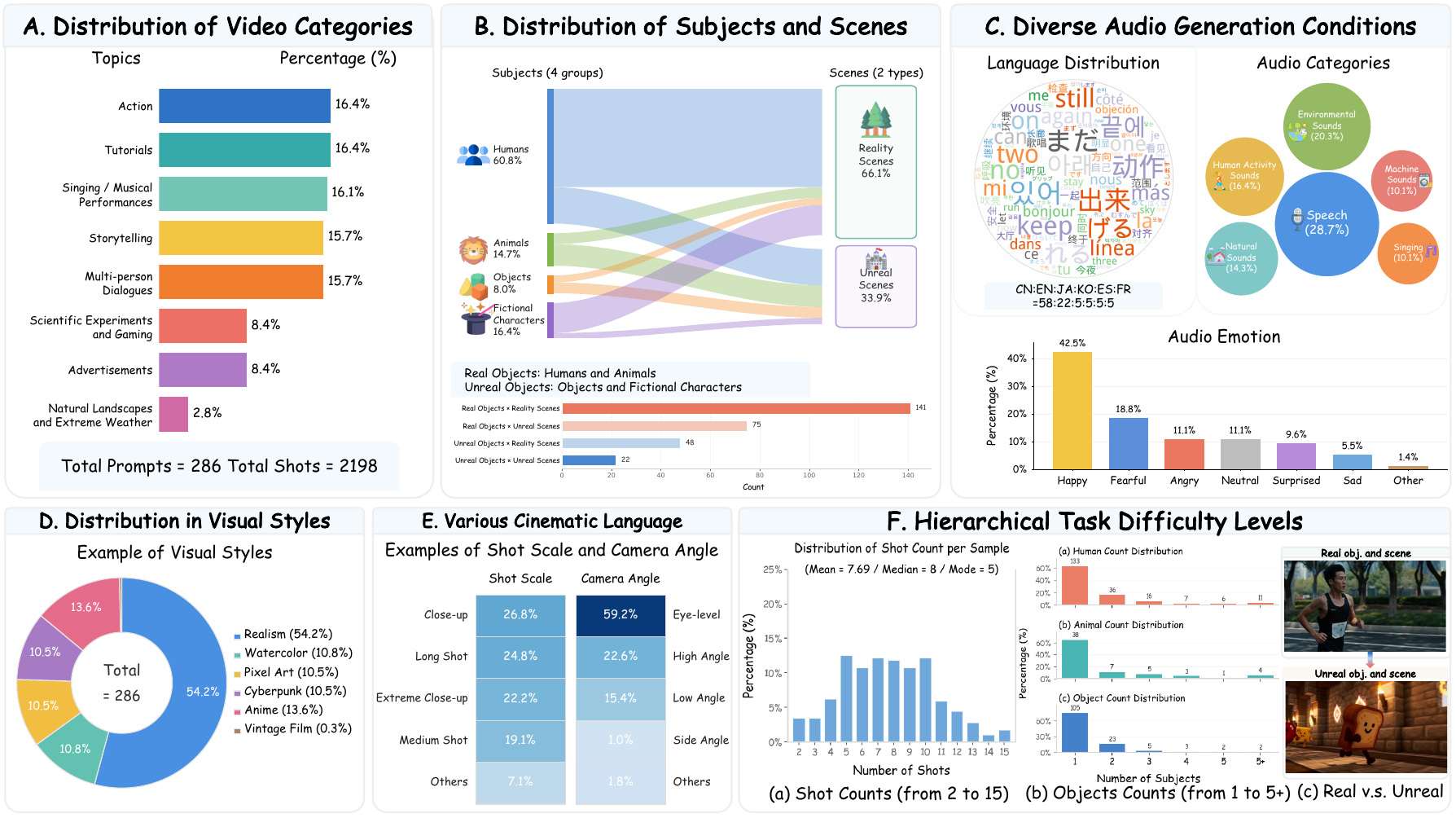}
    \caption{Diverse distribution of \benchnameplain. The benchmark covers diverse generation categories (A), realistic and non-realistic subjects and scenes (B), varied audio conditions and languages (C), diverse visual styles (D), rich cinematic requirements (E), and a broad range of task difficulty in terms of shot count, subject count, and scenario complexity (F). More statistics are in Appendix~\ref{app:data_analysis_summary}}
    \label{fig:analyse_main}
    \vspace{-3mm}
\end{figure}

\subsection{Data Analysis}
\label{sec:method:data-ana}

\textbf{Visual and stylistic diversity.}\quad
As detailed in \Cref{fig:analyse_main}(A) and (B), the benchmark balances 8 genres (\eg, Action) with demanding domains (\eg, Scientific Experiments). Subjects encompass 4 main categories (\eg, humans, fictional characters), situated across realistic (66.1\%) and non-realistic (33.9\%) scenes. Furthermore, \Cref{fig:analyse_main}(D) illustrates 6 diverse visual aesthetics; while realism dominates, multiple stylized domains (\eg, anime, cyberpunk) are included. This semantic and aesthetic diversity enables a comprehensive evaluation of models' adaptability and prompt adherence.

\textbf{Acoustic and linguistic diversity.}\quad
As illustrated in \Cref{fig:analyse_main}(C), our benchmark includes diverse audio content, emotional expressions, and languages. Audio conditions span 6 broad categories (\eg, speech and environmental noise), while explicitly annotated emotional attributes cover 7 distinct states (\eg, happiness and fear). Furthermore, spoken content is distributed across 6 languages to support rigorous evaluation of multilingual audio-visual alignment.

\textbf{Fine-grained cinematic language.}\quad
As shown in \Cref{fig:analyse_main}(E), we design professional cinematographic control into our benchmark. The prompts incorporate 5 major shot scales (\eg, close-up, long shot), 5 major camera angles, diverse camera movements (\eg, push-in, pan), and various lighting conditions. Additionally, we introduce multiple cross-shot transitions (\eg, hard cuts, fade-ins), facilitating a rigorous assessment of the cinematic generation capabilities of current models.

\textbf{Diverse reference assets.}\quad
To support reference-conditioned tasks (\eg, identity preservation and voice cloning), we provide 68 character images and 65 paired audio clips featuring extensive demographic and linguistic diversity. Additionally, 32 scene images across indoor and outdoor environments are included. These assets ensure robust conditioning for multi-modal generation.

\textbf{Multi-level task complexity.}\quad
As depicted in \Cref{fig:analyse_main}(F), we scale the shot count from 2 to 15, with an average of 7.7 shots per prompt. Beyond single-subject prompts, 32.2\% of prompts require multi-subject compositions, including scenarios with 5 or more simultaneous subjects. We further introduce challenging cases by cross-combining realistic and non-realistic subjects and scenes. This design facilitates systematic evaluation of models' capacities in long-form storytelling, complex spatial composition, and out-of-distribution generalization.

\subsection{Evaluation Suite}
\label{sec:method:evaluation_suite}

\subsubsection{Hierarchical Evaluation Metrics}
\label{sec:method:metrics}

We organize our evaluation metrics into four hierarchical levels, comprising 20 metrics in total (see~\Cref{fig:teaser}). More detailed descriptions of each metric are provided in Appendix~\ref{app:evaluation_suite_details}.

\textbf{Global-level metrics.}\quad
These metrics evaluate the overarching narrative, audio-visual alignment, and visual details across the entire video. 
\emph{1) Narrative coherence}: Assesses logical plot progression based on discrete events. 
\emph{2) Lip synchronization}: Evaluates lip-speech alignment across all dialogue shots. 
\emph{3) Sound attribution}: Measures the temporal overlap between visually active speakers and their audio. 
\emph{4) Audio-visual synchronization}: Measures the temporal offset between visual onsets and sound events. 
\emph{5) Visual quality}: Evaluates fine-grained visual fidelity.

\textbf{Cross-shot-level metrics.}\quad
These metrics assess the consistency of visual content, audio properties, and complex spatial layouts across consecutive shots. 
\emph{1) Cross-shot layout consistency}: Evaluates spatial layout coherence across shot transitions. 
\emph{2) Visual consistency}: A composite metric comprising five sub-metrics: consistency of subject, background, style, illumination, and color across shots. 
\emph{3) Music consistency}: Evaluates the stability of accompaniment, tempo, and rhythmic beats in non-speech background music across shots. 
\emph{4) Speaker timbre consistency}: Verifies that the distinct vocal identities of multiple speakers remain stable across different shots.

\textbf{Intra-shot-level metrics.}\quad
These metrics evaluate generation quality and prompt adherence within individual shots. 
\emph{1) Intra-shot layout-text alignment}: Assesses how accurately spatial layouts align with text prompts. 
\emph{2) Camera parameter adherence}: Evaluates compliance with the specified camera scale, angle, and movement. 
\emph{3) Audio quality}: Evaluates the acoustic quality of the generated audio. 
\emph{4) Text rendering accuracy}: Measures the correctness of visually rendered text. 
\emph{5) Word error rate}: Assesses speech transcription accuracy against the prompt-specified dialogue.

\textbf{Reference-level metrics.}\quad
These metrics assess fidelity to user-provided reference assets.
\emph{1) Subject fidelity}: Consistency with the reference image in appearance and identity.
\emph{2) Voice fidelity}: Consistency with the reference audio in vocal timbre.

\textbf{Overall score.}\quad
To avoid overemphasizing overlapping fine-grained aspects, we group related metrics into shared dimensions, merging five visual consistency metrics into \emph{Visual Quality} and four dialogue-related metrics into \emph{Multi-Speaker Dialogue Audio}, resulting in 11 final dimensions. We normalize these dimensions to $[0,1]$, average them, and multiply the result by a shot-completion penalty coefficient based on the ratio of generated shots to the specified shot count. As shown in Sec.~\ref{sec:exp:ablation}, this design aligns well with human expert judgments.

\begin{figure}[t]
    \centering
    \includegraphics[width=1\linewidth]{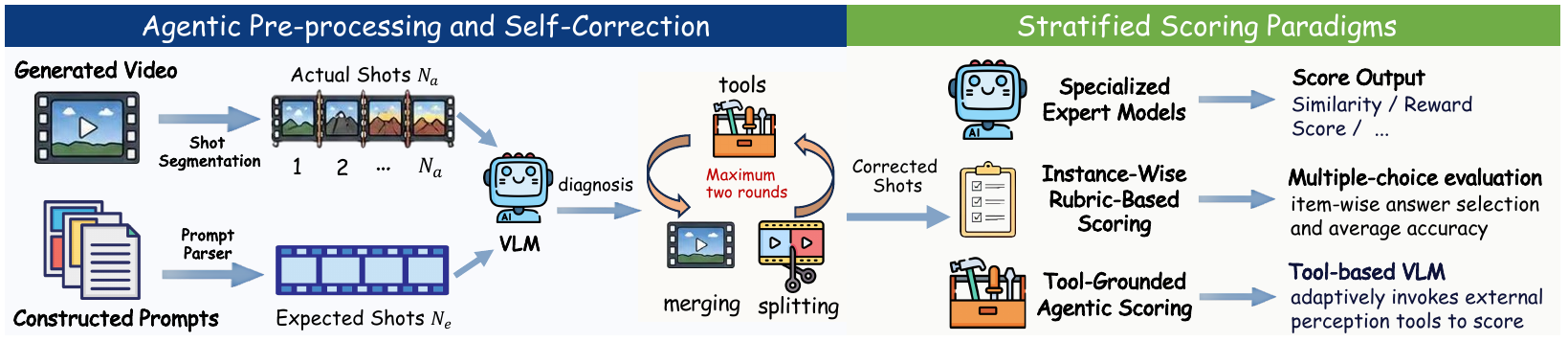}
    \caption{\textbf{Overview of the \benchnameplain evaluation framework.} We first perform agentic pre-processing with iterative shot self-correction to improve boundary quality. Metrics are then evaluated with stratified scoring paradigms, including expert models for well-defined tasks, rubric-based VLM scoring for subjective dimensions, and tool-grounded agentic scoring for complex properties.}
    \label{fig:eval_pipeline}
    \vspace{-3mm}
\end{figure}

\subsubsection{Adaptive Hybrid Evaluation Framework}
\label{sec:method:methodology}

Our evaluation framework consists of agentic self-correction and stratified scoring paradigms.

\textbf{Agentic pre-processing and self-correction.}\quad
To eliminate cascading failures caused by shot segmentation errors, we introduce an agentic pre-processing phase. Given a generated video, our framework first extracts initial temporal boundaries using TransNet V2~\citep{transnetv2}. Since direct boundary prediction by VLMs is unreliable, we employ a VLM (Qwen3.5~\citep{qwen35blog}) to iteratively inspect and evaluate the segments. The model determines whether specific shots require merging or splitting and invokes tools to refine the boundaries, thereby mitigating shot count anomalies. To balance accuracy and computational cost, we limit this process to a maximum of two iterations. In cases where the shot count remains mismatched after correction, the VLM performs a final shot-caption re-alignment, discarding non-aligned segments to ensure the integrity of downstream metric computations.

\textbf{Stratified scoring paradigms.}\quad
To balance evaluation cost, reliability, and comprehensiveness, we adopt three scoring paradigms based on metric complexity:
\emph{1) Specialized expert models}: For well-defined metrics (\eg, subject similarity), we use dedicated expert models for efficient evaluation while aligning with standard practice.
\emph{2) Instance-wise rubric-based scoring}: For subjective dimensions (\eg, narrative coherence), direct VLM scoring can be unstable. We therefore convert evaluation into prompt-specific rubrics, where the VLM answers predefined multiple-choice questions instead of producing unconstrained scalar scores.
\emph{3) Tool-grounded agentic scoring}: For complex compositional properties (\eg, layout-text consistency), pure VLM reasoning is often insufficient. 
We thus augment scoring by allowing models to adaptively invoke perception tools (\eg, object detectors and pose estimators) to extract objective evidence, which the VLM then uses to derive the final score.

\section{Experiments}
\label{sec:exp}

\subsection{Experimental Setup}
\label{sec:exp:setup}

We benchmark 19 representative video generators on \benchnameplain across two families.
\emph{(i) Closed-source commercial systems:} Seedance-2.0~\citep{seedance2.0}, Wan2.7-T2V~\citep{wan2.7}, Kling-V3-T2V~\citep{kling-3}, 
HappyHorse~\citep{happyhorse}, Sora-2~\citep{sora-2}, as well as reference-conditioned models Wan-R2V~\citep{wan2.7} and HappyHorse-R2V~\citep{happyhorse}.
\emph{(ii) Open-source pipelines:} We further divide them into five categories:
1) single-shot audio-video models that are concatenated shot by shot, including JavisDiT++~\citep{javisdit++}, JavisGPT~\citep{javisgpt}, MoVA~\citep{team2026mova} (TI2AV mode) and LTX-2.3~\citep{ltx-video} and daVinci-MagiHuman~\citep{magihuman} in T2AV and TI2AV modes;
2) multi-shot video models paired with dubbing models, such as ShotStream~\citep{shotstream} with HunyuanFoley;
3) single-shot video models that are dubbed and then concatenated shot by shot, such as Wan2.2~\citep{wan} with HunyuanFoley~\citep{hunyuanvideofoley} in TI2AV modes.
4) long-video generation models that can take multi-shot prompts as input and are paired with dubbing models, such as LongLive~\citep{longlive} with HunyuanFoley and Helios~\citep{helios} with HunyuanFoley;
5) Reference-Conditioned Models: DreamID-Omni~\citep{guo2026dreamid}.
Note that under the TI2AV setting, we utilize Wan2.7-Image~\citep{wan-image} to generate a storyboard (image set) from the scripts as multi-shot priors, with each image explicitly fed as the visual condition.

\subsection{Main Results}
\label{sec:exp:main}

\begin{table}[t]
  \centering
\caption{\textbf{Main results on \benchnameplain.}
The metrics are categorized into three dimensions: \textbf{Global:} \texttt{Narr.}: narrative coherence, \texttt{Lip}: lip synchronization, \texttt{Attr.}: sound attribution, \texttt{Sync}: audio-visual synchronization, \texttt{VQ}: visual quality. \textbf{Cross-Shot:} \texttt{C-Layout}: cross-shot layout consistency, \texttt{VC}: visual consistency, \texttt{Mus.}: music consistency, \texttt{Spk.}: speaker timbre consistency. \textbf{Intra-Shot:} \texttt{I-Layout}: intra-shot layout-text alignment, \texttt{Cam.}: camera parameter adherence, \texttt{PQ}: audio quality, \texttt{OCR}: text rendering accuracy, \texttt{WER}: word error rate. Top-5 cells per column are highlighted with a cyan gradient.}
  \label{tab:main-results}
  \scriptsize
  \setlength{\tabcolsep}{2.2pt}
  \renewcommand{\arraystretch}{1.05}
  \resizebox{\linewidth}{!}{%
  \begin{tabular}{l | ccccc | cccc | ccccc | c}
    \toprule
    \multirow{2}{*}{\textbf{Method}}
      & \multicolumn{5}{c|}{\textbf{Global}}
      & \multicolumn{4}{c|}{\textbf{Cross-Shot}}
      & \multicolumn{5}{c|}{\textbf{Intra-Shot}}
      & \multirow{2}{*}{\textbf{Overall}\,$\uparrow$} \\
      & \textbf{Narr.}\,$\uparrow$ & \textbf{Lip}\,$\uparrow$ & \textbf{Attr.}\,$\uparrow$ & \textbf{Sync}\,$\downarrow$ & \textbf{VQ}\,$\uparrow$
      & \textbf{C-Layout}\,$\uparrow$ & \textbf{VC}\,$\uparrow$ & \textbf{Mus.}\,$\uparrow$ & \textbf{Spk.}\,$\uparrow$
      & \textbf{I-Layout}\,$\uparrow$ & \textbf{Cam.}\,$\uparrow$ & \textbf{PQ}\,$\uparrow$ & \textbf{OCR}\,$\uparrow$ & \textbf{WER}\,$\downarrow$
      &  \\
    \midrule
    \rowcolor{gray!12}\multicolumn{16}{l}{\textit{Closed-source commercial systems}} \\
    Seedance-2.0~\citep{seedance2.0}                     & 0.816 & \rkfive{1.52} & 0.578 & 0.14 & \rkthree{0.795} & \rkone{0.809} & \rkfive{0.808} & 0.849 & 0.573 & \rkone{0.822} & 0.801 & \rkfour{6.51} & \rkone{0.726} & \rkfive{0.54} & \rkone{75.92} \\
    Wan2.7-T2V~\citep{wan2.7}                       & \rkfour{0.822} & 0.85 & \rkone{0.661} & 0.43 & \rkfive{0.773} & \rkfour{0.680} & 0.803 & \rkfour{0.880} & 0.641 & \rktwo{0.783} & 0.617 & 6.37 & 0.665 & \rkone{0.49} & \rkthree{72.26} \\
    Kling-V3-T2V~\citep{kling-3}                     & 0.796 & 1.02 & \rkfour{0.606} & 0.28 & \rktwo{0.801} & \rktwo{0.741} & \rktwo{0.856} & \rkthree{0.892} & \rkfour{0.657} & 0.609 & \rkfour{0.846} & 6.38 & 0.590 & 0.68 & \rkfour{72.25} \\
    HappyHorse~\citep{happyhorse}                       & \rkthree{0.825} & 0.73 & 0.579 & 0.24 & \rkone{0.804} & 0.632 & 0.790 & 0.833 & \rkthree{0.673} & 0.628 & 0.732 & \rkthree{6.60} & \rktwo{0.689} & \rkthree{0.51} & \rkfive{71.89} \\
    Sora-2~\citep{sora-2}                           & \rkone{0.852} & \rkthree{1.87} & 0.568 & 0.50 & \rkfour{0.792} & \rkthree{0.717} & \rkfive{0.808} & 0.834 & 0.520 & \rkfive{0.722} & 0.784 & 5.64 & \rkfive{0.675} & 0.75 & 71.19 \\
    \midrule
    \rowcolor{gray!12}\multicolumn{16}{l}{\textit{Open-source~\textcircled{1}: Native single-shot AV (concatenated shot-by-shot)}} \\
    LTX-2.3 (TI2AV)~\citep{ltx-video}               & 0.803 & 1.03 & 0.502 & \rktwo{0.07} & 0.732 & 0.670 & 0.762 & 0.767 & 0.522 & \rkthree{0.765} & \rkfive{0.814} & \rkone{6.96} & \rkthree{0.687} & \rktwo{0.49} & \rktwo{72.63} \\
    MoVA (TI2AV)~\citep{team2026mova}                     & \rktwo{0.839} & \rkfour{1.61} & 0.530 & \rkfive{0.12} & 0.681 & 0.626 & 0.790 & 0.801 & 0.496 & \rkfour{0.746} & 0.689 & \rkfive{6.40} & \rkfour{0.680} & 0.66 & 70.32 \\
    DaVinci+MagiHuman (TI2AV)~\citep{magihuman}        & 0.787 & \rktwo{3.08} & \rkfive{0.580} & \rkthree{0.07} & 0.685 & 0.422 & \rkfour{0.816} & \rktwo{0.957} & \rktwo{0.674} & 0.473 & 0.563 & 5.82 & 0.650 & 0.82 & 65.01 \\
    LTX-2.3 (T2AV)~\citep{ltx-video}                   & 0.768 & 0.96 & \rkthree{0.608} & \rkfour{0.09} & 0.754 & 0.439 & 0.596 & 0.770 & 0.562 & 0.348 & 0.781 & \rktwo{6.94} & 0.586 & \rkfour{0.53} & 64.40 \\
    DaVinci+MagiHuman (T2AV)~\citep{magihuman}         & 0.776 & \rkone{4.91} & \rktwo{0.654} & \rkone{0.05} & 0.699 & 0.267 & 0.586 & \rkone{0.958} & \rkone{0.699} & 0.494 & 0.472 & 5.78 & 0.164 & 0.83 & 60.65 \\
    JavisDiT++~\citep{javisdit++}                       & \rkfive{0.818} & 0.59 & 0.315 & 0.66 & 0.674 & 0.413 & 0.480 & 0.814 & 0.313 & 0.616 & 0.537 & 5.85 & 0.484 & 1.00 & 57.51 \\
    JavisGPT~\citep{javisgpt}                         & 0.745 & 0.42 & 0.113 & 0.54 & 0.633 & 0.351 & 0.554 & 0.792 & 0.097 & 0.362 & 0.624 & 6.09 & 0.268 & 0.99 & 53.95 \\
    \midrule
    \rowcolor{gray!12}\multicolumn{16}{l}{\textit{Open-source~\textcircled{2}: Long-video model + dubbing}} \\
    LongLive~\citep{longlive} + HunyuanFoley~\citep{hunyuanvideofoley}          & 0.783 & 0.70 & 0.284 & 0.40 & 0.703 & 0.589 & \rkone{0.857} & 0.830 & 0.261 & 0.289 & \rktwo{0.956} & 6.27 & 0.374 & 7.55 & 58.59 \\
    Helios~\citep{helios} + HunyuanFoley            & 0.748 & 0.68 & 0.138 & 0.79 & 0.685 & 0.583 & \rkthree{0.851} & 0.475 & \rkfive{0.646} & 0.151 & \rkthree{0.944} & 6.35 & 0.380 & 1.24 & 54.10 \\
    \midrule
    \rowcolor{gray!12}\multicolumn{16}{l}{\textit{Open-source~\textcircled{3}: Multi-shot video model + dubbing}} \\
    ShotStream~\citep{shotstream} + HunyuanFoley        & 0.782 & 1.03 & 0.543 & 0.41 & 0.677 & 0.280 & 0.748 & \rkfive{0.862} & 0.495 & 0.243 & 0.581 & 6.31 & 0.376 & 1.00 & 58.85 \\
    \midrule
    \rowcolor{gray!12}\multicolumn{16}{l}{\textit{Open-source~\textcircled{4}: Single-shot video-only model + dubbing (concatenated shot-by-shot)}} \\
    Wan2.2~\citep{wan} + HunyuanFoley (TI2AV)    & 0.794 & 1.19 & 0.378 & 0.43 & 0.685 & \rkfive{0.679} & 0.747 & 0.814 & 0.314 & 0.430 & \rkone{0.957} & 6.08 & 0.590 & 1.39 & 63.42 \\
    \bottomrule
  \end{tabular}%
  }
  \vspace{-3mm}
\end{table}

\Cref{tab:main-results} details the overall performance on \benchnameplain, from which we derive four key findings regarding current model bottlenecks.

\noindent\textbf{Finding~1: A significant performance gap persists between closed and open-source models, but modular agentic frameworks show potential to bridge it.}\quad 
Commercial systems (\eg, Seedance-2.0) consistently dominate the leaderboard. Native open-source multi-shot audio-video models remain absent, constrained by data scarcity and prohibitive computational costs. However, a modular ``image\,+\,audio-video'' pipeline decoupling per-shot keyframe synthesis from audio-video generation (\eg, LTX-2.3 in TI2AV mode) effectively boosts open-source performance to rival closed systems. This suggests that advancing beyond basic modularity toward \textit{a fully dynamic, agentic architecture may offer a viable, cost-effective path for the open community to challenge monolithic closed SOTA.}

\noindent\textbf{Finding~2: Compared to basic audio-visual fidelity, open-source models lag significantly behind closed systems in ``director-level'' structural control and cinematic language.}\quad 
Open-source models lag significantly behind closed systems in complex spatial and cinematic compliance. Markedly lower scores in layout alignment (\texttt{C-Layout}, \texttt{I-Layout}) and camera control (\texttt{Cam.}) suggest these models currently act as passive pixel renderers rather than fully controllable storytellers.

\noindent\textbf{Finding~3: Fine-grained joint audio-visual alignment remains an unsolved challenge for both closed and open-source models.}\quad 
Despite commendable unimodal generation quality, current systems still struggle with this inherent joint consistency. This is reflected in poor performance across lip-speech synchronization (\texttt{Lip}), sound attribution (\texttt{Attr.}), audio-visual synchronization (\texttt{Sync}), and multi-talker timbre consistency (\texttt{Spk.}). Accurately coupling phoneme-level audio with dynamic visual content across diverse cinematic languages remains a critical open problem.

\noindent\textbf{Finding~4: The alternative ``video-first, post-hoc dubbing'' paradigm is inadequate for complex multi-shot audio-video generation.}\quad 
Relying on independent models (\eg, HunyuanFoley) to dub pre-generated videos causes severe speech distortion (high \texttt{WER}) and poor lip-sync. This occurs because post-hoc dubbing lacks frame-level semantic grounding across hard camera cuts, disrupting joint cross-modal alignment. Conversely, unified architectures are essential for the MSAV task.

\subsection{Performance Analysis on Complex Scenarios}
\label{sec:exp:difficulty}

\begin{figure}[t]
    \centering
    \includegraphics[width=1\linewidth]{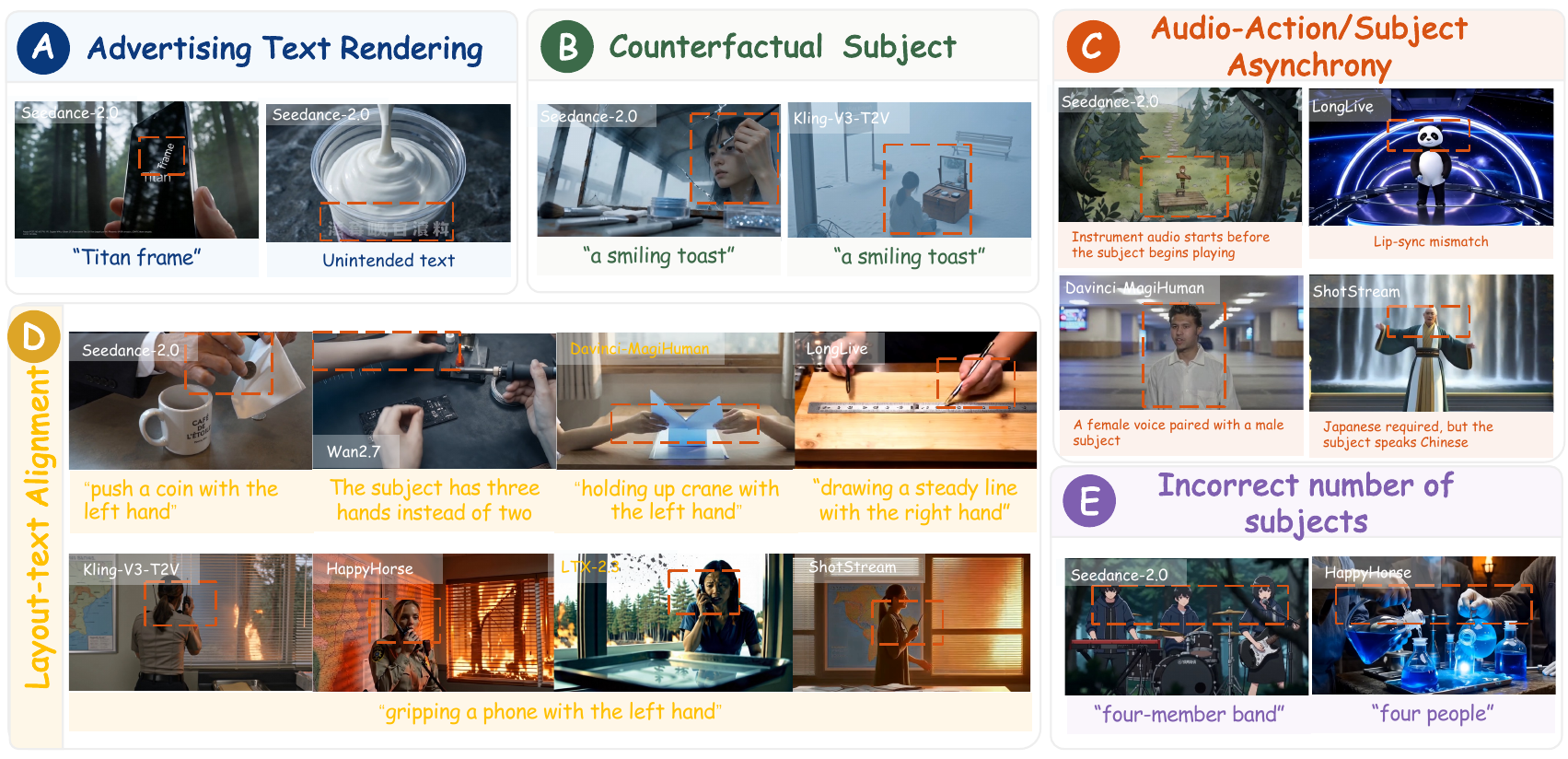}
    \captionsetup{skip=0pt}
    \caption{\textbf{Qualitative failure cases of evaluated models.} Examples include text rendering errors (A), counterfactual subject mismatches (B), audio-visual synchronization failures (C), layout control failures (D), and incorrect subject counts (E).}
    \label{fig:bad_case}
    \vspace{-3mm}
\end{figure}

\noindent\textbf{Qualitative analysis on challenging cases.}\quad
As illustrated in \Cref{fig:bad_case}, the evaluated models exhibit five recurring failure modes.
\emph{1) Text rendering failures}: even leading closed-source models, such as Seedance-2.0~\citep{seedance2.0}, still struggle with fine-grained text generation, often producing misspelled or unintended text.
\emph{2) Counterfactual subject failures}: models may fail to generate subjects that match the prompt, such as producing an incorrect object instead of \texttt{a smiling toast}.
\emph{3) Audio-visual synchronization failures}: common issues include lip-sync mismatch and audio-action asynchrony even in top closed-source systems, while open-source models, such as Davinci-MagiHuman~\citep{magihuman} and ShotStream~\citep{shotstream}, often show more severe errors, including audio-subject mismatch and failure to generate speech in the required language.
\emph{4) Layout control failures}: for prompts with spatial constraints, such as left-right hand relations, both closed-source models (\eg, Seedance-2.0) and open-source models (\eg, LTX) often fail to satisfy the required body-part configuration.
\emph{5) Subject count failures}: models frequently generate the wrong number of subjects in complex multi-subject scenes.
These cases highlight that robust MSAV generation still requires substantial progress in controllability, compositionality, and audio-visual alignment.

\begin{figure}[t]
  \begin{minipage}[t]{0.48\linewidth}
    \centering
\captionof{table}{\textbf{Overall score across prompts with different required shot counts.}}
\vspace{-1mm}
\setlength{\tabcolsep}{3.5pt}
\renewcommand{\arraystretch}{1.05}
\resizebox{\linewidth}{!}{%
\begin{tabular}{lccc}
\toprule
& \multicolumn{3}{c}{\textbf{Shot Count Range}} \\
\cmidrule(lr){2-4}
\textbf{Method} & \textbf{1--4} & \textbf{5--10} & \textbf{11--15} \\
\midrule
Seedance-2.0                     & 77.70 & 75.30 & 76.00 \\
Wan2.7-T2V                       & 73.50 & 71.70 & 72.10 \\
Kling-V3-T2V                     & 73.50 & 72.30 & 70.00 \\
HappyHorse                       & 74.90 & 70.60 & 72.80 \\
LTX-2.3 (TI2AV)                  & 75.10 & 72.00 & 72.50 \\
DaVinci+MagiHuman (TI2AV)        & 70.90 & 64.70 & 62.30 \\
Wan2.2 + HunyuanFoley (TI2AV)    & 71.80 & 62.80 & 60.10 \\
ShotStream + HunyuanFoley        & 59.60 & 58.20 & 59.70 \\
LongLive + HunyuanFoley          & 66.10 & 61.10 & 41.60 \\
JavisDiT++                       & 60.40 & 58.50 & 59.30 \\
\bottomrule
\end{tabular}%
}
\label{tab:diff:shotnum}

  \end{minipage}\hfill
  \begin{minipage}[t]{0.48\linewidth}
    \centering
\captionof{table}{\textbf{Overall score across realistic and non-realistic prompts.}}
\vspace{-1mm}
\setlength{\tabcolsep}{4pt}
\renewcommand{\arraystretch}{1.05}
\resizebox{\linewidth}{!}{%
\begin{tabular}{lcc}
\toprule
\textbf{Method} & \textbf{Real.} & \textbf{Non-Real.} \\
\midrule
Seedance-2.0                     & 76.80 & 74.50 \\
Wan2.7-T2V                       & 73.40 & 70.50 \\
Kling-V3-T2V                     & 73.50 & 70.30 \\
HappyHorse                       & 72.50 & 70.50 \\
LTX-2.3 (TI2AV)                  & 74.20 & 70.50 \\
DaVinci+MagiHuman (TI2AV)        & 66.10 & 63.40 \\
Wan2.2 + HunyuanFoley (TI2AV)    & 64.20 & 61.90 \\
ShotStream + HunyuanFoley        & 59.90 & 56.70 \\
LongLive + HunyuanFoley          & 60.30 & 56.00 \\
JavisDiT++                       & 61.00 & 56.40 \\
\bottomrule
\end{tabular}%
}
\label{tab:diff:realistic}

  \end{minipage}\hfill
  \vspace{-5mm}
\end{figure}

\noindent\textbf{Quantitative analysis on shot counts and realistic vs.\ non-realistic data.}\quad 
We identify two main performance bottlenecks. \textbf{1) Shot count:} As shown in \Cref{tab:diff:shotnum}, performance declines across all models when the required shot count increases from $1$--$4$ to $5$ and beyond, demonstrating the inherent difficulty of long-horizon generation. Notably, open-source models degrade significantly more. For example, from $1$--$4$ to $11$--$15$ shots, closed-source Kling-V3-T2V drops by $3.5\%$, whereas open-source LongLive$+$HunyuanFoley collapses by $24.5\%$ and Wan2.2$+$HunyuanFoley by $11.7\%$. This highlights multi-shot consistency as a critical weakness for open-source pipelines. \textbf{2) Realistic vs.\ non-realistic data:} As illustrated in \Cref{tab:diff:realistic}, overall scores noticeably decrease on non-realistic prompts on all methods. Closed-source models like Seedance-2.0 drop by $2.3\%$, while open-source models face steeper declines (\eg, JavisDiT++ drops by $4.6\%$). This indicates that generating out-of-distribution visual contents universally compromises performance across all model families.

\begin{wraptable}{r}{0.51\textwidth}
  \vspace{-4mm}
  \centering
  \caption{\textbf{Results on reference-to-AV generation.}
  }
  \vspace{-2mm}
  \setlength{\tabcolsep}{4pt}
  \renewcommand{\arraystretch}{1.05}
  \small
  \resizebox{\linewidth}{!}{%
  \begin{tabular}{lccc}
  \toprule
  \textbf{Method} & \textbf{Img-DINO}$\uparrow$ & \textbf{Img-Face}$\uparrow$ & \textbf{Voice}$\uparrow$ \\
  \midrule
  Wan-R2V~\citep{wan2.7}               & \underline{0.208} & \textbf{0.368}    & \textbf{0.657} \\
  HappyHorse-R2V~\citep{happyhorse}        & \textbf{0.259}    & \underline{0.244} & \underline{0.545} \\
  DreamID-Omni~\citep{guo2026dreamid}  & 0.119             & 0.054             & 0.535 \\
  \bottomrule
  \end{tabular}%
  }
  \label{tab:diff:reference}
  \vspace{-4mm}
\end{wraptable}

\noindent\textbf{Quantitative analysis on reference-conditioned generation.}\quad
\Cref{tab:diff:reference} reveals a substantial visual fidelity gap between open- and closed-source models. 
Open-source DreamID-Omni significantly trails Wan-R2V and HappyHorse-R2V on Img-DINO and Img-Face, yet its voice similarity ($0.535$) closely approaches the closed-source HappyHorse-R2V ($0.545$). This highlights that visual preservation is harder than voice cloning in joint audio-visual customization, making cross-modal fidelity balance a critical direction for future research.

\subsection{Human Preference Alignment and Evaluation Robustness}
\label{sec:exp:ablation}

To validate the reliability of our benchmark, we measure alignment with human judgments and robustness across different VLM judges (see Appendix~\ref{app:human_annotation} for annotation details).

\begin{wraptable}{r}{0.5\textwidth}
  \vspace{-0.8em}
  \centering
  \setlength{\tabcolsep}{4pt}
  \renewcommand{\arraystretch}{1.05}
  \small
  \resizebox{\linewidth}{!}{%
  \begin{tabular}{llc}
  \toprule
  \textbf{Metric} & \textbf{Method} & \textbf{Spearman} \textbf{$\rho_s\uparrow$} \\
  \midrule
  \textbf{Overall} 
    & \textbf{Ours} & \textbf{0.915} \\
  \midrule
  \multirow{3}{*}{\makecell[l]{Narrative \\ Coherence}}
    & Direct VLM Scoring (Qwen3.5) & 0.600 \\
    & Instance-wise Rubric (Qwen2.5-VL) & 0.820 \\
    & \textbf{Instance-wise Rubric (Qwen3.5)} & \textbf{0.850} \\
  \midrule
  \multirow{3}{*}{\makecell[l]{Cross-Shot Layout \\ Consistency}}
    & Direct VLM Scoring (Qwen3.5) & 0.429 \\
    & Tool-Grounded (Qwen2.5-VL) & 0.732 \\
    & \textbf{Tool-Grounded (Qwen3.5)} & \textbf{0.767} \\
  \midrule
  \multirow{3}{*}{\makecell[l]{Intra-Shot Text-Layout \\ Alignment}}
    & Direct VLM Scoring (Qwen3.5) & 0.405 \\
    & Tool-Grounded (Qwen2.5-VL) & 0.741 \\
    & \textbf{Tool-Grounded (Qwen3.5)} & \textbf{0.786} \\
  \bottomrule
  \end{tabular}%
  }
  \vspace{-0.4em}
  \caption{\textbf{Agreement with human experts.} Our overall ranking and metric designs show strong correlation with human judgments and remain robust across different foundation models.}
  \label{tab:abl:human_robustness}
  \vspace{-1.0em}
\end{wraptable}

\noindent\textbf{Alignment with human perception.}\quad We employ Spearman's ($\rho_s$) rank correlation~\citep{spearman1961proof} to measure consistency with expert human ratings. \textit{1) Overall ranking:} As shown in \Cref{tab:abl:human_robustness}, 
our overall score achieves a high $\rho_s$ of $0.915$, confirming strong alignment with human judgments. 
\textit{2) Complex metrics:} We validate the reliability of our metric designs over direct VLM scoring on three challenging metrics. For narrative coherence, cross-shot layout consistency, and intra-shot layout consistency, our instance-wise rubrics and tool-grounded agentic scoring improve Spearman correlation by $0.250$, $0.338$, and $0.381$, reaching $\rho_s=0.850$, $0.767$, and $0.786$, respectively. These results show that rubric-based decomposition and tool-grounded evidence are critical for aligning automatic evaluation with human judgment on complex tasks.

\noindent\textbf{Robustness across VLM backbones.}\quad
We further substitute the underlying judge from Qwen3.5~\citep{qwen35blog} to the smaller Qwen2.5-VL-32B-Instruct~\citep{qwen3-vl} to assess robustness.
As reported in \Cref{tab:abl:human_robustness}, our rubric- and tool-grounded designs remain highly stable across backbones (\eg, dropping only slightly from $0.850$ to $0.820$ on narrative coherence), and still vastly outperform direct VLM scoring. This demonstrates that \benchnameplain's evaluation framework is robust to the specific VLM choice, further validating the reliability of our metric design.

\section{Conclusion}
\label{sec:conclusion}

We present \benchname, the first multi-shot audio-video generation benchmark with an adaptive hybrid evaluation framework. Our benchmark provides comprehensive coverage of data dimensions and challenging scenarios, including video, audio, shot, and reference aspects, and supports reliable evaluation through agentic shot self-correction and stratified scoring paradigms. 
Our evaluation of 19 state-of-the-art systems shows that modular and agentic open-source pipelines have the potential to narrow the gap with closed-source models. However, current models still remain far from director-level generation, particularly in cinematic control and fine-grained audio-visual synchronization. We believe that \benchnameplain, together with the insights it provides, will serve as a rigorous benchmark and diagnostic tool for future MSAV research.



\bibliographystyle{plain}
\bibliography{secs/6_references}

\appendix
\clearpage

\begin{center}
  \Large\textbf{Appendix}
\end{center}
\startcontents
\printcontents{}{1}{\setcounter{tocdepth}{2}}
\clearpage

\section{More Data Details on \benchnameplain}
\label{app:more_details_msavbench}

\subsection{Data Design Details}
\label{app:data_design_details}

\benchnameplain organises every prompt along the four orthogonal data-design dimensions of Sec.~\ref{sec:method:data-def} (Video, Audio, Shot, Reference). Each dimension is annotated with a set of sub-attributes. Subjects and scenes are independently classified into two top-level reality classes -- \emph{realistic} and \emph{non-realistic} -- where the non-realistic class encompasses both \emph{coherent fictional} (e.g.\ cyberpunk city) and \emph{counterfactual} (e.g.\ a frozen tropical desert) sub-types.

\textbf{Dim.~1 -- Video.} Four sub-attributes:
(i) \emph{video genre} ($8$ categories): Action, Narrative, Tutorial, Singing\,\&\,Music performance, Multi-person Dialogue, Science / Game, Advertising, Nature;
(ii) \emph{visual style} ($6$ styles): photo-realistic, anime, watercolour storybook, pixel art, cyberpunk, retro film;
(iii) \emph{subject type} ($4$ classes): humans, animals, inanimate objects, fictional characters;
(iv) \emph{scene type}: realistic and non-realistic.

\textbf{Dim.~2 -- Audio.} Three sub-attributes:
(i) \emph{audio content class} ($6$ categories): speech, singing, instrument / machine, human-made environment (e.g.\ laughter, footsteps), natural ambient, mixed (foley with voice-over, music with environment);
(ii) \emph{audio emotion} ($7$ emotions): joy, fear, anger, surprise, sadness, neutrality, awe;
(iii) \emph{spoken language} ($6$ values): Chinese, English, Japanese, Korean, Spanish, French.

\textbf{Dim.~3 -- Shot (cinematic language).} Five sub-attributes annotated per shot:
(i) \emph{shot scale} ($5$ types): close-up, mid-close, mid, mid-long, long;
(ii) \emph{shot angle} ($5$ types): eye-level, top-down, low-angle, oblique, dutch;
(iii) \emph{camera motion} ($4$ types): push-pull, pan-tilt, tracking, hand-held/shake;
(iv) \emph{transition} ($4$ types): hard cut, dissolve, match cut, fade;
(v) \emph{lighting} ($5$ types): natural, side, soft, neon, low-key.

\textbf{Dim.~4 -- Reference.} Three sub-attributes that are paired with a prompt: (i) \emph{subject reference image} ($68$ images); (ii) \emph{paired reference audio} ($65$ audio clips paired with the subject images); (iii) \emph{scene reference image} ($32$ indoor / outdoor environments). All reference assets are assigned across $96$ prompts.

\subsection{Data Construction Details}
\label{app:data_construction_details}

\subsubsection{Expert-Curated Sub-Category Vocabulary}
\label{app:subcategories}

The seed taxonomy used in Stage~1 of data construction contains an $8$-genre top level whose second-level vocabulary on the released suite totals $144$ fine-grained sub-categories.

The eight top-level genres and a representative subset of each genre's fine-grained sub-categories are:
\textbf{(C1) Action} ($32$ sub-categories): martial-arts duel, kungfu choreography, weapon combat, parkour, street-dance battle, ballet, modern dance, basketball, football, swimming, gymnastics, boxing, skateboarding, rock climbing, surfing, marathon, BASE jumping, bungee jumping, wingsuit flying, card shuffling, surgical suturing.
\textbf{(C2) Narrative storytelling} ($19$ sub-categories): detective reasoning, family warmth, romance, sci-fi adventure, historical legend, comedy, horror/thriller, coming-of-age, workplace drama, road-trip movie, human-animal interaction, animal documentary, fantasy adventure, war epic, courtroom drama.
\textbf{(C3) Tutorial} ($14$ sub-categories): cooking, building blocks, origami, painting tutorial, instrument-fingering tutorial, fitness routine, makeup tutorial, woodworking, electronic soldering, gardening, CPR demonstration.
\textbf{(C4) Singing \& music performance} ($17$ sub-categories): solo pop / rock / folk / classical / rap, choir, band performance, conductor, piano solo, guitar solo, violin solo, drums solo, guzheng, saxophone, orchestral ensemble, street performance, music festival.
\textbf{(C5) Multi-person dialogue} ($18$ sub-categories): family dinner, street encounter, classroom discussion, hospital visit, in-transit conversation, courtroom debate, news interview, talk-show panel, elevator small-talk, whispered exchange, negotiation, casual gossip.
\textbf{(C6) Scientific experiment / game} ($16$ sub-categories): chemistry experiments (acid-base reaction, crystallisation, combustion, colour change), biology experiments (microscopy, plant growth, dissection), physics experiments (optical refraction, electromagnetic induction, fluid dynamics, free fall), astronomy observation, electronic games, board games, sports games.
\textbf{(C7) Advertising} ($19$ sub-categories): sneaker, smartphone, automobile, perfume, food and beverage, skincare, movie trailer, game trailer, sports event, tourism destination, app UI demo, e-commerce listing.
\textbf{(C8) Nature \& extreme weather} ($\sim$$9$ sub-categories): aurora, volcanic eruption, deep-sea bioluminescence, forest fire, super-cell thunderstorm, sand storm, glacier collapse, polar night, monsoon rainfall.

\begin{figure}[t]
    \centering
    \includegraphics[width=1\linewidth]{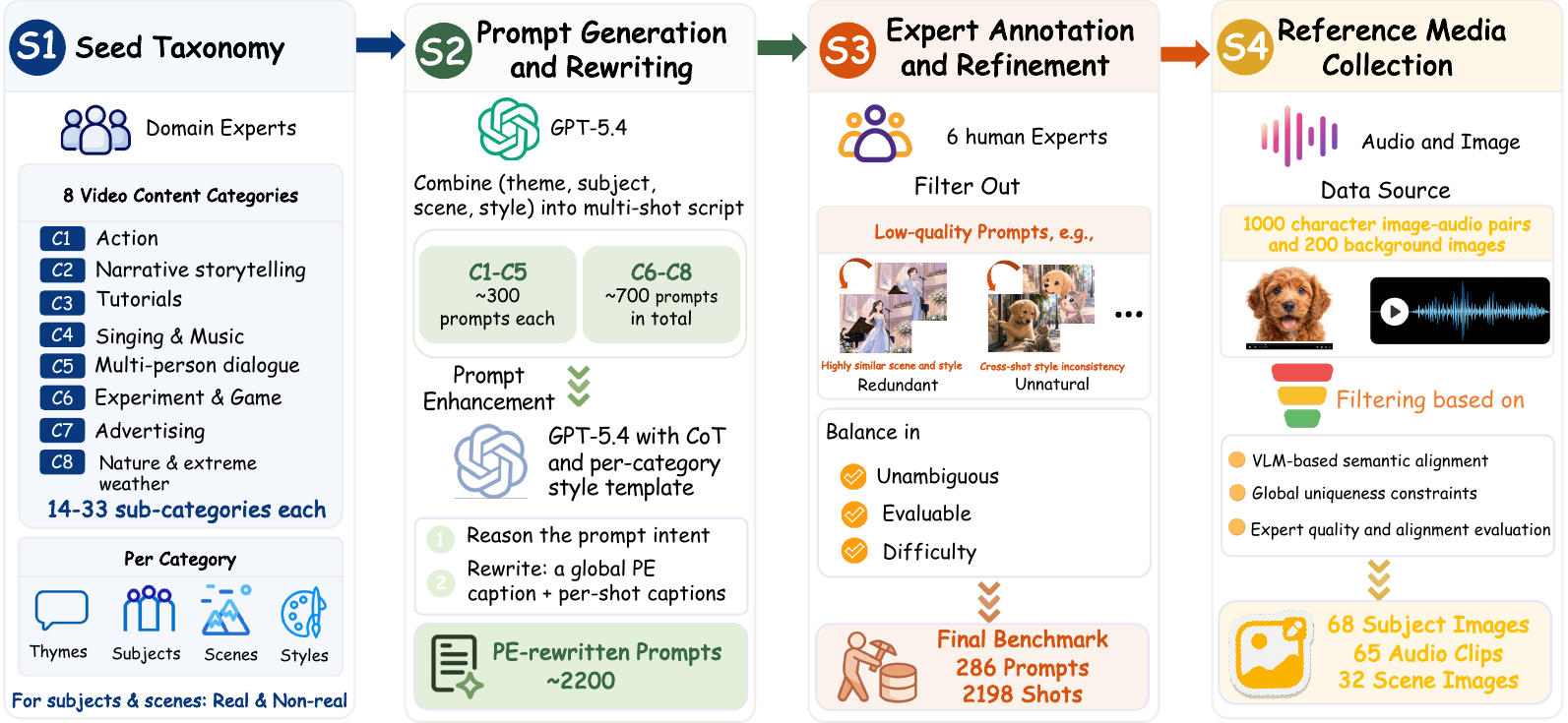}
    \caption{\textbf{The data construction pipeline of \benchnameplain.}
    (1) Domain experts define an eight-category seed taxonomy with fine-grained sub-categories, with diverse types of subject, scene, and visual style.
    (2) GPT-5.4 first samples $(theme, subject, scene, style)$ quadruples and synthesises an initial multi-shot script with structured per-shot metadata; a Prompt-Enhancement (PE) model then rewrites it into the global-to-shot format with explicit cinematic language.
    (3) Six domain experts review every PE-rewritten script, filter out low-quality / hallucinated cases, and refine ambiguous descriptions.
    (4) Reference media are sampled from public benchmarks, automatically tagged by Gemini~3.1~Pro, and finally curated by experts to obtain a clean reference-conditioned subset.}
    \label{fig:data_construct}
\end{figure}

\subsubsection{LLM Prompt Templates}
\label{app:prompt_templates}

Stage~2 of data construction relies on two GPT-5.4 system templates: an \emph{initial-prompt} template that turns a sampled $(theme, subject, scene, style)$ quadruple plus a target shot count into a structured multi-shot script with all evaluation metadata, and a \emph{Prompt-Enhancement (PE)} template that rewrites the initial script into the cinematic global-to-shot format consumed by downstream generators. We show the two templates below.

\begin{tcolorbox}[
  enhanced,breakable,
  colback=gray!4,colframe=black!60,
  arc=2pt,boxrule=0.5pt,
  fonttitle=\bfseries\small,
  title={Initial-Prompt System Template (GPT-5.4)}
]
\scriptsize\ttfamily
You are an expert prompt designer for multi-shot audio-video generation. Given the dimension constraints below and a target shot count, write one high-quality multi-shot prompt and emit the structured metadata in JSON.\\[3pt]
\textbf{[Constraints]}\\
\textbf{(1) Anchor category} (one of $8$): the narrative core must revolve around it.\\
\textbf{(2) Global attributes} (constant across the whole video, choose one value each): visual style $\in$ \{photo-realistic, anime, watercolour, pixel, cyberpunk, retro film\}; subject; colour tone (warm / cold / neutral); audio content (speech / singing / instrument / human-made env. / natural ambient / mixed).\\
\textbf{(3) Per-shot attributes} (may vary, but with continuity): \emph{scene} (transitions narratively coherent, no abrupt jumps; do not freeze on one scene); \emph{lighting} ($1$--$2$ dominant variants); \emph{audio emotion} ($1$--$2$ dominant tones); \emph{spoken language} (single language, no code-switching); \emph{shot scale / angle / camera motion / transition} (must vary for cinematic richness).\\
\textbf{(4) Cross-call diversity.} Vary style / subject / scene across calls. Single- or multi-character setups allowed. \textbf{Never use real personal names}; use descriptive references (``a young woman'').\\
\textbf{(5) Continuity.} Shots must follow a clear causal / temporal order, each a natural continuation of the previous one.\\
\textbf{(6) Per-category guidance.} Apply the $8$-category guidance (Action / Narrative / Tutorial / Singing\,\&\,Music / Multi-person Dialogue / Science\,\&\,Game / Advertising / Nature).\\[3pt]
\textbf{[Writing rules]} produce exactly the requested shot count; every shot must be concrete and visually evocative; output one continuous paragraph (\emph{global framing} $\to$ \emph{Shot 1} $\to$ \emph{Shot 2} $\to \dots$); no bullet points; output language matches the user input.\\[3pt]
\textbf{[Output format -- JSON only]}\\
\{ "prompt": "<the multi-shot prompt>",\\
\hspace*{1em}"dimensions": \{\\
\hspace*{2em}"video\_content","video\_style","video\_subject","video\_tone","audio\_content": "<chosen value>",\\
\hspace*{2em}"num\_persons","num\_animals","num\_object\_subjects": <int>,\\
\hspace*{2em}"per\_shot": [ \{ "shot\_id":1, "scene","lighting","audio\_emotion","language",\\
\hspace*{4em}"shot\_scale","shot\_angle","camera\_motion","transition": "<value>" \}, \dots ] \} \}\\
Output JSON only; no extra text.
\end{tcolorbox}

\begin{tcolorbox}[
  enhanced,breakable,
  colback=gray!4,colframe=black!60,
  arc=2pt,boxrule=0.5pt,
  fonttitle=\bfseries\small,
  title={Prompt-Enhancement (PE) System Template (GPT-5.4)}
]
\scriptsize\ttfamily
You are an expert prompt engineer for multi-shot AI video generation. Given a user request, return one JSON object:\\[2pt]
\{ "is\_safe": true|false, "reasoning": "...", "caption": "...",\\
\hspace*{1em}"category": "Action | Narrative | Tutorial | Singing\&Music | Multi-person Dialogue | Science\&Game | Advertising | Nature" \}\\[3pt]
\textbf{[Safety]} \texttt{is\_safe=false} only on sexual / graphically violent / politically sensitive content; otherwise true.\\[3pt]
\textbf{[Reasoning rules]} Produce a structured, decision-oriented analysis (directional, no visual expansion):
(i) pick the top-level category from one of the $8$ above;
(ii) identify the core technical challenge (e.g.\ lip-sync, dense text rendering, multi-shot editing logic);
(iii) plan the shot count ($1$--$15$) and the role of each shot (establishing, close-up emotion, action climax);
(iv) plan cinematic language: per-shot scale (long / full / medium / close / extreme close), angle (low / eye-level / high), camera motion (push-in / pull-out / pan / tilt), with explicit cross-shot diversity;
(v) plan a unified colour tone and a $1$--$2$-style lighting plan;
(vi) plan audio: content type, $1$--$2$ dominant emotional tones, a single spoken language, AV-sync points.
Apply per-category guidance from the given list.\\[3pt]
\textbf{[Caption rules]} Rewrite the user input into a single multi-shot caption with the following template:\\
$\bullet$ \textbf{Global framing} -- open with ``\emph{This video contains $X$ shots; it is a $\langle$category$\rangle$ in $\langle$style$\rangle$, $\langle$colour tone$\rangle$, accompanied by $\langle$audio content$\rangle$ in $\langle$language$\rangle$.}'' Lighting and emotional colour are anchored on $1$--$2$ dominant variants.\\
$\bullet$ \textbf{Per-shot detail} -- each shot begins with ``\emph{Shot $N$ [\,$t_s$--$t_e$\,s\,]}'' (default total duration $15$\,s) and describes, in order: transition from previous shot; shot scale and angle (diverse across shots); composition and focus; scene; lighting; subject state and props; subject motion / physical change; audio emotional tone, language, and the precise AV interaction; camera-motion trajectory.\\
$\bullet$ \textbf{Continuity} -- each shot is a natural continuation of the previous one; respect temporal and causal order; no abrupt jumps.\\
$\bullet$ \textbf{Naming} -- replace any real personal name in the input with a descriptive reference (``a young woman'').\\[3pt]
\textbf{[Output format]} a single coherent natural-language paragraph (\emph{global framing} $\to$ \emph{Shot 1} $\to$ \emph{Shot 2} $\to \dots$); no bullet points, no extra headings; rich vocabulary; output language matches the user input.\\[3pt]
Output JSON only; no extra text.\\
\textbf{[User original prompt]}:
\end{tcolorbox}

\subsection{Data Analysis Details}
\label{app:data_analysis_summary}

\begin{figure}[t]
    \centering
    \begin{minipage}{0.49\linewidth}\centering
        \includegraphics[width=\linewidth]{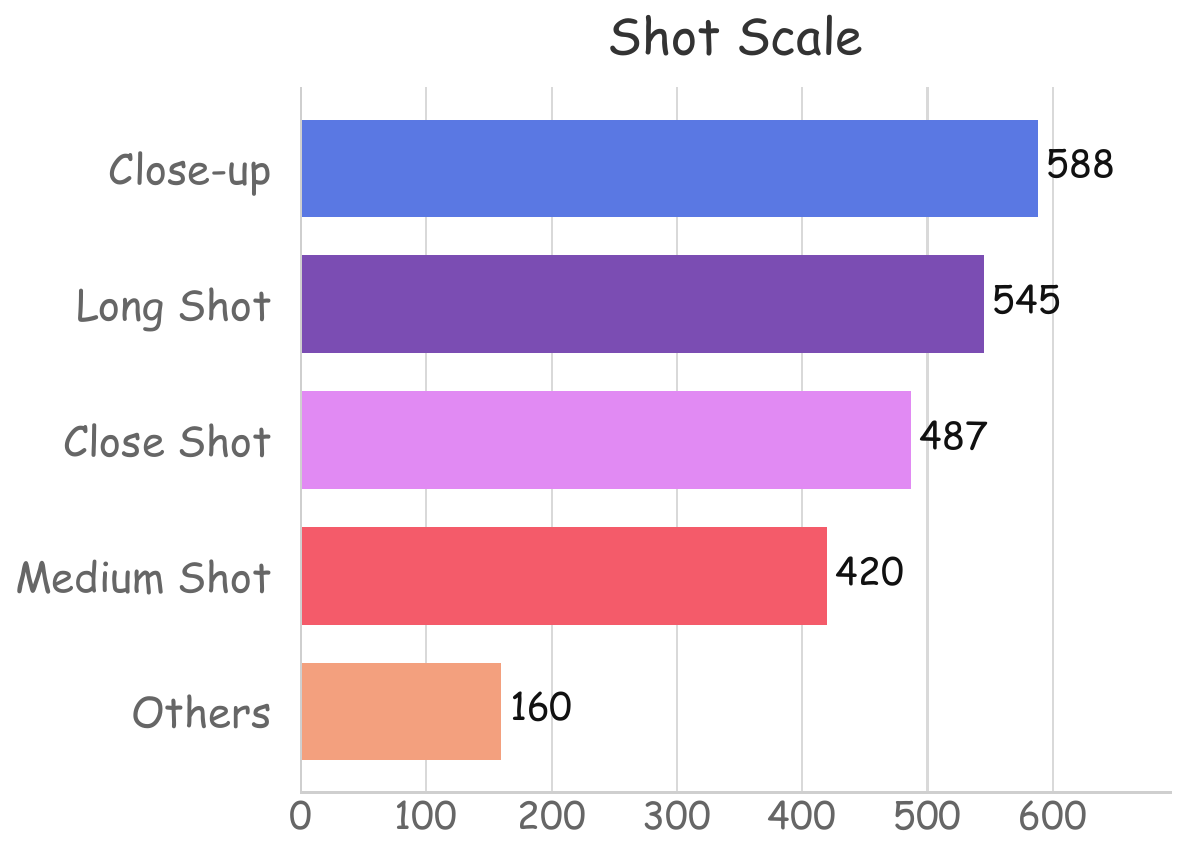}\\
        \scriptsize (a) Shot scale distribution (top-$4$ + tail).
    \end{minipage}\hfill
    \begin{minipage}{0.49\linewidth}\centering
        \includegraphics[width=\linewidth]{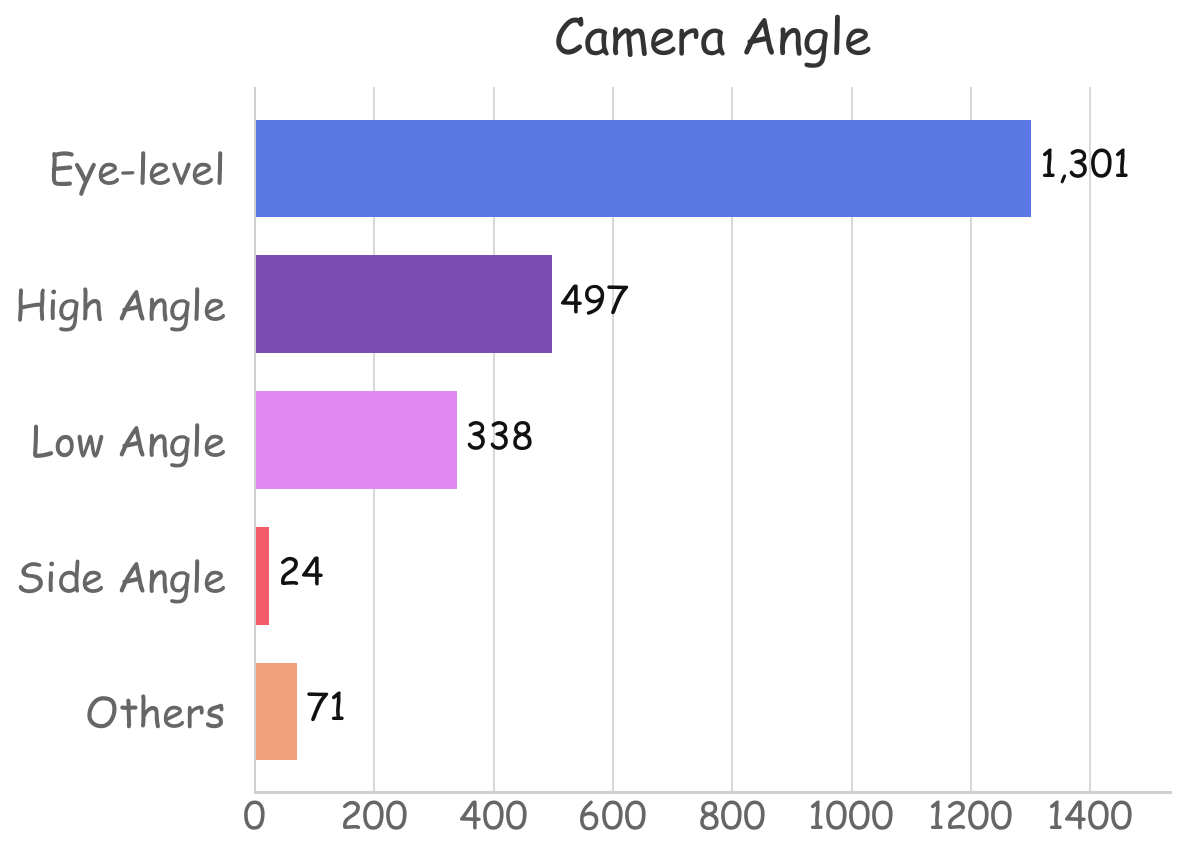}\\
        \scriptsize (b) Camera angle distribution (top-$4$ + tail).
    \end{minipage}\\[6pt]
    \begin{minipage}{0.49\linewidth}\centering
        \includegraphics[width=\linewidth]{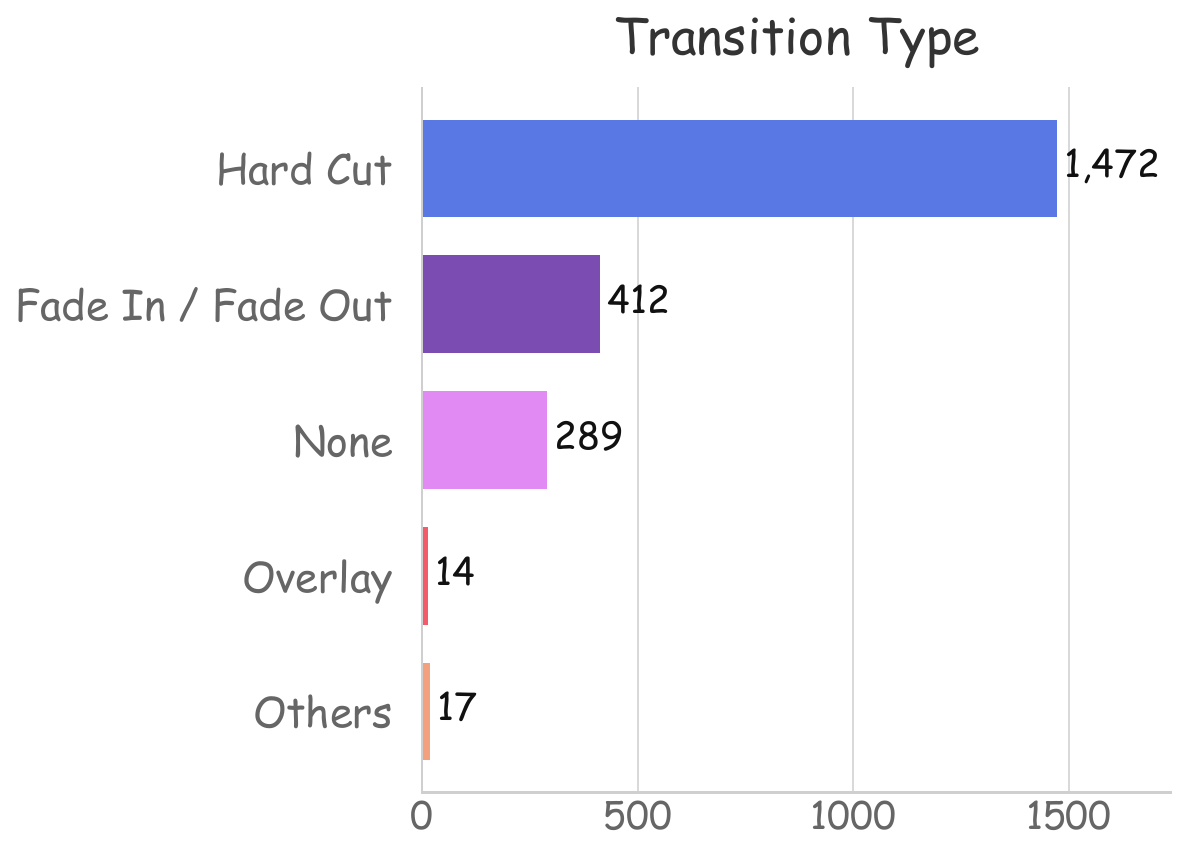}\\
        \scriptsize (c) Transition-type distribution (top-$4$ + tail).
    \end{minipage}\hfill
    \begin{minipage}{0.49\linewidth}\centering
        \includegraphics[width=\linewidth]{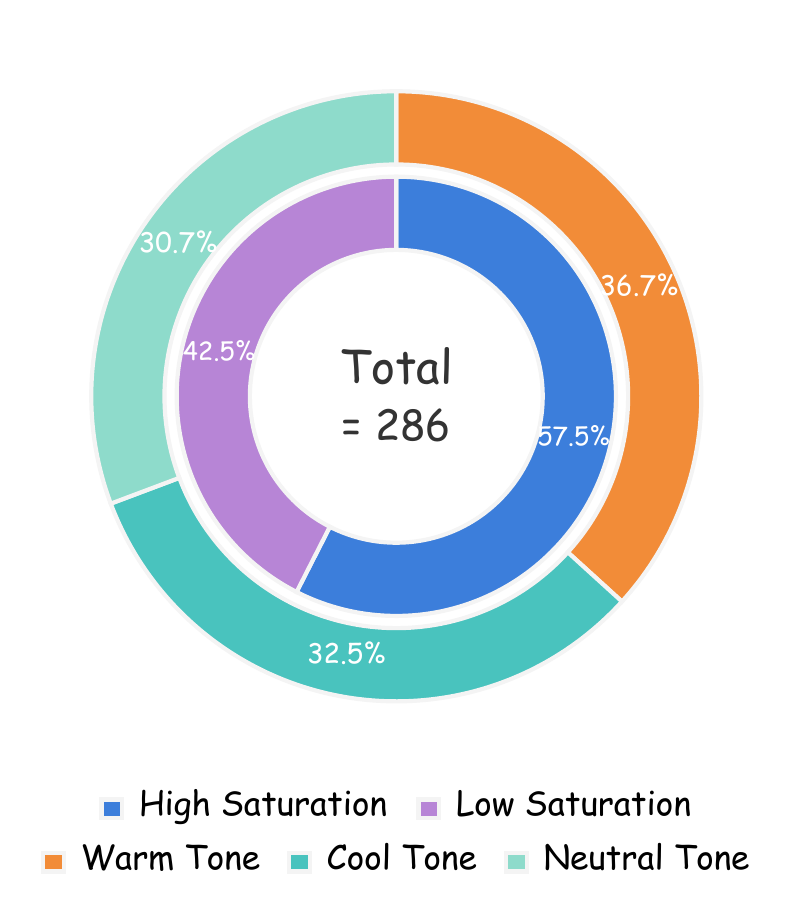}\\
        \scriptsize (d) Colour tone $\times$ saturation (prompt-level).
    \end{minipage}
    \caption{\textbf{Long-tail cinematic-language and tonal distributions of \benchnameplain.} Shot scale, camera angle, transition type and tone$\times$saturation distributions on the released $286$-prompt suite.
    }
    \label{fig:more_data_distribution}
\end{figure}

We summarise the released $286$-prompt / $2{,}198$-shot benchmark below. The high-level distributions are visualised in \Cref{fig:analyse_main} (in the main text), and the cinematic-language distributions (shot scale, camera angle, transition, tone $\times$ saturation) are reported in \Cref{fig:more_data_distribution}.

\textbf{Visual and stylistic diversity.}
The eight video genres are balanced: Action $16.4\%$, Tutorial $16.4\%$, Narrative $15.7\%$, Singing\,\&\,Music $16.1\%$, Multi-person Dialogue $15.7\%$, Science\,/\,Game $8.4\%$, Advertising $8.4\%$, Nature $2.8\%$.
Subjects span humans ($60.8\%$), animals ($14.7\%$), inanimate objects ($8.0\%$) and fictional characters ($16.4\%$); scenes are realistic ($66.1\%$) versus non-realistic ($33.9\%$).
Six visual styles are represented at the prompt level: photo-realistic $54.2\%$, anime $13.6\%$, watercolour $10.8\%$, pixel art $10.5\%$, cyberpunk $10.5\%$, retro film $0.3\%$.

\textbf{Acoustic and linguistic diversity.}
Audio content per prompt is dominated by speech ($28.7\%$), human-made environmental sounds ($20.3\%$), nature ambient ($14.3\%$), instrument / machine ($10.1\%$), singing ($10.1\%$), and human activity sounds ($16.4\%$).
Per-shot emotional colour spans the seven categories: joy ($42.5\%$), fear / suspense ($18.8\%$), anger / tension ($11.1\%$), neutral ($11.1\%$), surprise ($9.6\%$), sad ($5.5\%$), and others ($1.4\%$).
Spoken content is distributed across six languages -- Chinese $165$ prompts, English $64$, Japanese $15$, Korean $15$, Spanish $14$, French $13$ -- enabling explicit multilingual evaluation.

\textbf{Cinematic language.}
\benchnameplain reports $5$ major shot scales (close-up $26.8\%$, long $24.8\%$, extreme close $22.2\%$, mid $19.1\%$, mid-close+mid-long $5.2\%$, plus a $1.9\%$ tail) and $5$ major shot angles (eye-level $59.2\%$, top-down $22.6\%$, low $15.4\%$, side $1.0\%$, others $1.8\%$). 
Camera motion is reported as $4$ major types (push-pull $44.6\%$, pan-tilt $26.5\%$, tracking-and-orbit $5.8\%$, hand-held / shake $23.1\%$); transitions span $4$ major types (hard cut $66.9\%$, dissolve $18.7\%$, none $13.0\%$, match cut / fade $1.4\%$); and lighting is reported with $5$ major types (natural, side, soft, neon, low-key).
The distributions are plotted in \Cref{fig:more_data_distribution}.

\textbf{Reference assets.}
The released reference subset contains $68$ subject reference images, $65$ paired reference audio clips and $32$ scene reference images, all assigned across $96$ prompts.
Subjects span both realistic and anime domains; reference audio clips cover five age buckets ($0$--$19$, $19$--$30$, $31$--$45$, $46$--$60$, $60{+}$), multiple ethnicities and six languages;
scene reference images cover both indoor environments (restaurants, bedrooms, offices) and outdoor environments (snack streets, playgrounds, parks, coastal areas, courtyards, grasslands).

\textbf{Multi-level task complexity.}
Shot count per prompt ranges from $2$ to $15$ (mean $7.7$): $7\%$ have $2$--$3$ shots, $19\%$ have $4$--$5$, $23\%$ have $6$--$7$, $23\%$ have $8$--$9$, $18\%$ have $10$--$11$, $7\%$ have $12$--$13$ and $3\%$ have $14$--$15$. $32.2\%$ of prompts require multi-subject composition, with over $10\%$ demanding $\geq 5$ simultaneous subjects. Cross-combining reality classes yields four difficulty buckets: realistic-subject $\times$ realistic-scene $49.3\%$, realistic-subject $\times$ non-realistic-scene $26.2\%$, non-realistic-subject $\times$ realistic-scene $16.8\%$, non-realistic-subject $\times$ non-realistic-scene $7.7\%$.

\section{More Evaluation Suite Details on \benchnameplain}
\label{app:evaluation_suite_details}

\subsection{Metric Definitions, Tools and Score Mapping}
\label{app:metric_table}

Our evaluation framework contains 20 metrics organized into four levels: \textit{Story}, \textit{Cross-Shot}, \textit{Intra-Shot}, and \textit{Reference}.
For each metric, we briefly specify \emph{(i)} what it measures, \emph{(ii)} the tool or judge used, \emph{(iii)} how it is computed, and \emph{(iv)} how the raw output is mapped to a score in $[0,1]$.

\subsubsection{Story-Level Metrics}
\label{app:metric_story}

\textbf{(1) Narrative coherence.}
Measures whether the video forms a coherent story or valid procedural sequence across shots.
It is evaluated by a rubric-based VLM judge (Qwen 3.5~\citep{qwen35blog}) over uniformly sampled frames from the full video.
The judge answers predefined binary questions about event ordering, causal validity, and completeness.
The final score is the proportion of positive answers.

\textbf{(2) Visual quality.}
Measures whether prompt-specified visual attributes are correctly realized.
It is evaluated by a rubric VLM judge using prompt-instantiated multiple-choice questions.
Each prompt slot is converted into an MCQ and scored by answer accuracy.
The final score is the average MCQ accuracy.

\textbf{(3) Audio-visual synchronization.}
Measures temporal synchronization between visual events and sound at the whole-video level.
It is evaluated by DeSync metric, which is predicted by the Synchformer model~\citep{synchformer}.
The tool predicts the global audio-video offset.
The raw offset $\Delta t$ is first mapped to $[0,1]$ by $\max(0, 1-|\Delta t|/2.0\text{\,s})$.

\textbf{(4) Lip-speech synchronization.}
Measures lip-sync quality for dialogue-bearing shots.
It is evaluated using active-speaker localization~\citep{lr-asd}, speaker diarization~\citep{sortformer}, and StableSyncNett~\citep{latentsync}.
Matched speaking segments are scored and averaged across the video.
The raw sync confidence is directly used as the score.

\textbf{(5) Sound attribution.}
Measures whether speech is temporally aligned with the correct visible speaker.
It is evaluated using visual active-speaker detection~\citep{lr-asd} and audio diarization~\citep{sortformer}.
Speakers are matched across modalities and their temporal overlap is computed.
The final score is the mean overlap ratio.

\subsubsection{Cross-Shot-Level Metrics}
\label{app:metric_crossshot}

\textbf{(6) Cross-shot layout consistency.}
Measures spatial coherence of the main subject across shots, including position, orientation, scale, and prompt-specified hand relations.
It is evaluated by a tool-grounded agentic judge with grounding~\citep{grounding} and pose~\citep{bazarevsky2020blazepose} tools.
The score is computed from adjacent-shot consistency checks.
The final score is the average pass rate.

\textbf{(7) Subject consistency.}
Measures identity and appearance consistency of the main subject across shots.
It is evaluated using a VLM localizer~\citep{qwen35blog}, DINOv2\citep{oquab2023dinov2}, and ArcFace~\citep{deng2019arcface}.
Subject crops are extracted and encoded, and pairwise similarities are computed across shots.
The final score is $\max(0,\cos)$ averaged over pairs.

\textbf{(8) Background consistency.}
Measures background stability across shots after removing the foreground subject.
It is evaluated using foreground erasure and CLIP~\citep{clip} image embeddings.
Background embeddings are compared pairwise across shots.
The final score is the mean clipped cosine similarity.

\textbf{(9) Style consistency.}
Measures whether the visual style remains consistent across shots.
It is evaluated using CSD-ViT-L~\citep{somepalli2024measuring} style embeddings.
Pairwise cosine similarities are computed across shots.
The final score is the mean clipped cosine similarity.

\textbf{(10) Illumination consistency.}
Measures stability of lighting, shadow, and brightness across shots.
It is evaluated by a rubric-based VLM judge over sampled frames.
Adjacent shot pairs are checked for lighting consistency.
The final score is the average pass rate.

\textbf{(11) Colour consistency.}
Measures consistency of tone, saturation, and contrast across shots.
It is evaluated by a rubric-based VLM judge.
Adjacent shot pairs are compared for color consistency.
The final score is the average pass rate.

\textbf{(12) Music consistency.}
Measures continuity of background music across shots.
It is evaluated using Demucs~\citep{defossez2019demucs}, MuQ~\citep{defossez2019demucs}, and MIR-AIDJ All-in-onee~\citep{taejun2023allinone}.
The score combines music embedding similarity, BPM agreement, and beat alignment.
The final score is a weighted sum of these components in $[0,1]$.

\textbf{(13) Voice timbre consistency.}
Measures speaker timbre consistency across dialogue-bearing shots.
It is evaluated using VAD~\citep{SileroVAD}, Demucs~\citep{defossez2019demucs}, and w2v-BERT-2.0~\citep{chung2021w2v} speaker embeddings.
Per-shot speaker embeddings are extracted and compared across shots.
The final score is the mean clipped cosine similarity.

\subsubsection{Intra-Shot-Level Metrics}
\label{app:metric_intrashot}

\textbf{(14) Intra-shot layout-text alignment.}
Measures whether the spatial arrangement and hand actions within a shot match the shot caption.
It is evaluated by a tool-grounded agentic VLM judge with grounding and pose tools.
The judge answers predefined sub-questions for each shot.
The final score is the average pass rate.

\textbf{(15) Camera parameter adherence.}
Measures adherence to prompt-specified shot scale, angle, motion, and framing.
It is evaluated by a rubric VLM judge over sampled shot frames.
Each specified camera attribute is checked independently.
The final score is computed as the average pass rate.

\textbf{(16) Audio quality.}
Measures acoustic and production quality of generated audio.
It is evaluated using Audiobox-Aesthetic~\citep{tjandra2025aes}.
We use its production-quality sub-score for each shot.
The raw score is mapped to $[0,1]$ by $(\mathrm{PQ}-1)/9$.

\textbf{(17) Text rendering accuracy.}
Measures character-level fidelity of rendered on-screen text.
It is evaluated using PP-OCRv5~\citep{cui2025paddleocr30technicalreport,cui2025paddleocrvlboostingmultilingualdocument,cui2026paddleocrvl15multitask09bvlm} on advertising-style prompts.
Recognized text is compared against the target text using character error rate.
The final score is $1-\mathrm{CER}$, clipped to $[0,1]$.

\textbf{(18) ASR transcription (WER).}
Measures speech transcription accuracy against the prompt-specified script.
It is evaluated using FireRedASR2-LLM~\citep{xu2026fireredasr2s} or Whisper-large-v3~\citep{radford2023robust}, depending on language.
The transcription is compared against the target using word error rate.
The final score is $1-\min(\mathrm{WER},1)$.

\subsubsection{Reference-Level Metrics}
\label{app:metric_reference}

\textbf{(19) Subject fidelity.}
Measures whether the generated subject matches the reference image in identity and appearance.
It is evaluated using the same subject-embedding pipeline as cross-shot subject consistency.
Generated subject embeddings are compared with the reference image embedding.
The final score is the mean clipped cosine similarity.

\textbf{(20) Voice fidelity.}
Measures whether the generated speaker matches the reference voice in timbre.
It is evaluated using the same speaker-embedding pipeline as cross-shot voice consistency.
Generated speech embeddings are compared with the reference voice embedding.
The final score is the mean clipped cosine similarity.

\subsubsection{Overall Score Aggregation}
\label{app:overall_score}

Some atomic metrics reflect fine-grained sub-dimensions of the same underlying capability and therefore partially overlap in evaluation scope. Treating them as separate dimensions would over-weight that capability in the final aggregation.
We thus merge five visual consistency metrics into \emph{Visual Quality} and four dialogue-related audio metrics into \emph{Multi-Speaker Dialogue Audio}, yielding 11 final dimensions. 
Specifically, subject, background, style, illumination, and color consistencies are combined into \emph{Visual Quality}, while voice timbre consistency, lip-sync, sound attribution, and ASR transcription are combined into \emph{Multi-Speaker Dialogue Audio}. The remaining dimensions are kept separate.

All dimensions are mapped to $[0,1]$ using metric-specific deterministic rules as described above, and then averaged. To account for structural failures in multi-shot generation, we further multiply the average by a shot-completion penalty coefficient, defined as the proportion of valid generated shots relative to the prompt-specified shot count. As reported in \Cref{tab:abl:human_robustness}, this design shows strong alignment with human judgments.

\subsection{Stratified Scoring Paradigms}
\label{app:scoring_paradigms}

All 20 metrics are implemented with one of three scoring paradigms.

\paragraph{(1) Specialized expert models ($10$ metrics).}
These metrics are computed by task-specific expert models or deterministic signal-processing pipelines without VLM-based reasoning.
The corresponding metrics are:
{audio-visual sync.},
{lip-speech sync.},
{sound attribution},
{style consistency},
{music consistency},
{voice timbre consistency},
{audio quality},
{text rendering accuracy},
{ASR transcription (WER)}, and
{voice fidelity}.

\paragraph{(2) Instance-wise rubric-based scoring ($5$ metrics).}
These metrics are computed by a single-pass VLM judge using fixed rubrics, with the final score given by the pass rate over applicable sub-questions.
The corresponding metrics are:
{narrative coherence},
{visual quality},
{illumination consistency},
{colour consistency}, and
{camera parameter adherence}.

\paragraph{(3) Tool-grounded agentic scoring ($5$ metrics).}
These metrics rely on tool-grounded evaluation, where localized evidence from perception tools is used to support scoring.
The corresponding metrics are:
{cross-shot layout consistency},
{subject consistency},
{background consistency},
{intra-shot layout-text alignment}, and
{subject fidelity}.

\section{Additional Experimental Details}
\label{app:experimental_details}

\subsection{Implementation}
\label{app:implementation}

All perception tools are deployed as independent FastAPI micro-services on $8\!\times\!\text{A100}$ hosts. GPT-5.4~\citep{gpt-5.4} is used for initial prompt generation and prompt enhancement. For VLM-based evaluation, Gemini~3.1~Pro~\citep{gemini-3.1-pro} is used for audio-related judgments, while Qwen3.5~\citep{qwen35blog} is used for visual-related judgments. Tool outputs are cached at the case level and reused across metrics whenever possible.

\subsection{Cost-Efficient Evaluation}
\label{app:cost_reuse}

Evaluating multi-shot audio-video generation is inherently challenging, requiring a careful balance between evaluation accuracy and computational cost. Our framework is designed to remain efficient in both tool usage and VLM calls. \textit{First}, not all metrics rely on VLM judges: many metrics are handled by specialized expert models or deterministic pipelines, which substantially reduces evaluation cost. \textit{Second}, intermediate results are reused across metrics whenever possible; for example, shared outputs from subject localization, embedding extraction, foreground removal, OCR, and ASR are computed once and consumed by multiple metrics. \textit{Third}, our framework is robust to different VLM backbones, and the smaller Qwen-based judge still achieves competitive alignment with human judgments, as shown in Sec.~\ref{sec:exp:ablation}.

\section{Human Expert Annotation}
\label{app:human_annotation}

\subsection{Experts for Benchmark Construction}
\label{app:construction_experts}

The Stage~1 taxonomy design and Stage~3 prompt curation in data construction pipeline are carried out by six domain experts, all of whom are full-time researchers in AIGC and audio-video generation. Each expert holds a graduate degree in computer vision, multimedia, or audio signal processing.
During Stage~3, each PE-rewritten prompt is reviewed by at least two experts; disagreements on filtering or refinement are escalated to a third senior expert and resolved by majority vote.
After this process, $286$ of the original $2{,}200$ PE-rewritten prompts ($13.0\%$) are retained in the released benchmark, highlighting the strictness of the curation process.

\subsection{Evaluation Experts and Pairwise Annotation Protocol}
\label{app:evaluation_experts}

For the human-alignment study in Sec.~\ref{sec:exp:ablation}, we recruit two groups of annotators, all of whom are full-time AIGC researchers and aesthetic-quality annotators with prior experience in aesthetic-quality annotation. The first group consists of $30$ experts for system-level evaluation, comparing $16$ video-generation models in terms of overall quality. Each annotator labels $40$ video pairs, yielding a total of $1{,}200$ pairwise judgments. The second group consists of $10$ experts for fine-grained evaluation on three metrics: narrative coherence, cross-shot layout consistency, and intra-shot layout-text alignment. For each metric, annotators compare $10$ candidate methods and label $36$ pairs each, resulting in $360$ judgments per metric. To ensure broad coverage, video pairs are uniformly sampled across genres, including realistic and stylized content, single- and multi-subject scenes, and videos with varying numbers of shots.

To reduce annotation bias, all videos are anonymized and presented in random order, and annotators follow a unified rubric for each evaluation metric. They are allowed to select one of three outcomes for each pair: ``A wins,'' ``B wins,'' or ``both good / both bad.'' Ties are counted as $0.5$ for each method when computing win rates. The resulting method rankings are then compared with automatic metrics using Spearman's $\rho$.

\subsection{Annotation Interface}
\label{app:annotation_interface}

\begin{figure}[t]
    \centering
    \includegraphics[width=0.8\linewidth]{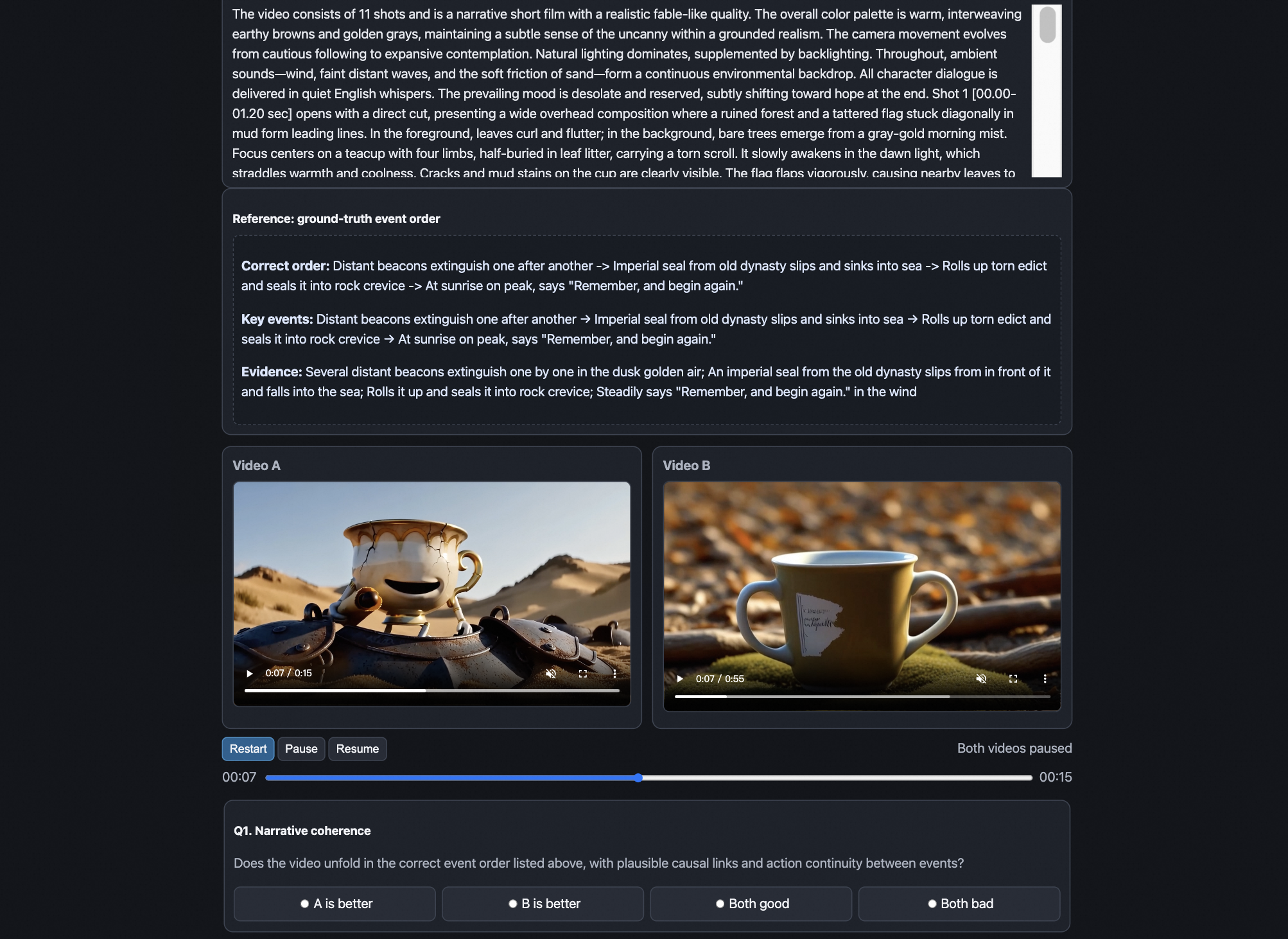}
    \caption{Screenshot of the annotation interface used for pairwise expert evaluation.}
    \label{fig:annotation_interface}
\end{figure}

Human evaluation is conducted via a custom web interface for fully anonymized pairwise comparison. Annotators are presented with two candidate videos together with the corresponding prompt and relevant metadata, and select the preferred result under the specified evaluation criterion. The resulting pairwise preferences are aggregated into system-level rankings. Figure~\ref{fig:annotation_interface} illustrates the interface.

\section{Ethics, Privacy, and Licensing}
\label{app:ethics}

The text prompts in \benchnameplain are synthetically generated from expert-designed taxonomies and subsequently reviewed by domain experts. They do not contain personal data, identifiable individuals, sensitive political or geographic content, or real proper names. The reference images and audio clips are drawn from previously published benchmarks with open redistribution terms and are used in accordance with their respective licenses. We further review these assets to exclude content that may raise privacy or cultural-sensitivity concerns.

Accordingly, \benchnameplain does not introduce new privacy risks. The generated videos used in our experiments are produced solely for evaluation and are not redistributed. Upon release, we will provide the prompt set, the reference assets that can be legally shared, and the evaluation framework.

\section{Limitations}
\label{app:limitations}

We discuss the limitations of our \benchnameplain.
First, some components of our agentic evaluation pipeline rely on multimodal foundation models as judges, which may introduce additional cost in large-scale evaluations.
Nevertheless, as shown in Sec.~\ref{sec:exp:ablation}, the framework remains well aligned with human judgment even when instantiated with a smaller open-source model, suggesting that our evaluation method is robust to the choice of VLM backbone.
Second, because there is not yet a mature open-source model that natively supports multi-shot audio-video generation, some of our baseline constructions follow a staged generation paradigm built on top of existing model capabilities. As more native joint audio-video multi-shot generation models become available, they can be incorporated into \benchnameplain for a more comprehensive evaluation.


\end{document}